\documentclass[12pt]{article}
\usepackage{amsmath}
\usepackage{graphicx,psfrag,epsf}
\usepackage{enumerate}
\usepackage{natbib}
\usepackage{url} 
\usepackage{hyperref, bm}
\usepackage{algpseudocode}
\usepackage{algorithm}
\usepackage{amsmath,amssymb}
\usepackage{bbm}
\usepackage{xcolor}
\usepackage{epstopdf}
\usepackage{caption}
\usepackage{array}
\usepackage{multirow}
\usepackage{amsfonts,amstext,amsmath,amssymb,amsthm}
\usepackage{mathtools}
\DeclarePairedDelimiter{\floor}{\lfloor}{\rfloor}
\DeclarePairedDelimiter{\ceil}{\lceil}{\rceil}

\newcolumntype{P}[1]{>{\centering\arraybackslash}p{#1}}

\newcommand{\bbf}{\bm{f}}
\newcommand{\bbR}{\mathbb{R}}

\newtheorem{definition}{Definition}

\newtheorem{lemma}{Lemma}
\newtheorem{example}{Example}

\newtheorem{theorem}{Theorem}

\newcommand{\argmin}{\mathop{\mathrm{argmin}}}

\newcommand\numberthis{\addtocounter{equation}{1}\tag{\theequation}}

\newcommand{\blind}{0}
\begin{document}

\def\spacingset#1{\renewcommand{\baselinestretch}%
{#1}\small\normalsize} \spacingset{1}


\if0\blind
{
  \title{\bf Calibrating multi-dimensional complex ODE from noisy data via deep neural networks}
  \author{Kexuan Li\\
    Global Analytics and Data Sciences, Biogen Inc\\
    Fangfang Wang \\
    Department of Mathematical Science, Worcester Polytechnic Institute\\
    Ruiqi Liu\\
    Department of Mathematics and Statistics, Texas Tech University\\
    Fan Yang\\
    Eli Lilly and Company\\
    Zuofeng Shang\\
    Department of Mathematical Sciences, New Jersey Institute of Technology
}
  \maketitle
} \fi

\if1\blind
{
  \bigskip
  \bigskip
  \bigskip
  \begin{center}
    {\LARGE\bf Title}
\end{center}
  \medskip
} \fi

\bigskip
\begin{abstract}
Ordinary differential equations (ODEs) are widely used to model complex dynamics that arise in biology, chemistry, engineering, finance, physics, etc. Calibration of a complicated ODE system using noisy data is generally challenging. In this paper, we propose a two-stage nonparametric approach to address this problem. We first extract the de-noised data and their higher order derivatives using boundary kernel method, and then feed them into a sparsely connected deep neural network with rectified linear unit (ReLU) activation function. Our method is able to recover the ODE system without being subject to the curse of dimensionality and the complexity of the ODE structure. We have shown that our method is consistent if the ODE possesses a general modular structure with each modular component involving only a few input variables, and the network architecture is properly chosen. Theoretical properties are corroborated by an extensive simulation study that also demonstrates the effectiveness of the proposed method in finite samples. Finally, we use our method to simultaneously characterize the growth rate of COVID-19  cases from the 50 states of the  United States.
\end{abstract}

\noindent%

\spacingset{1.3}
\newpage
\section{Introduction}
The use of ordinary differential equations (ODEs) is prevalent in both social  and natural sciences to study complex dynamic  phenomena or dynamical systems. For instance, linear ODEs are often used to describe population growth (\cite{IntroPopulation}), Lorenz equation---a high-dimensional nonlinear ODE system---widely used to characterize chaos systems (\cite{IntroLorenz}), and high-dimensional linear ODEs used to construct a dynamic gene regulatory network (\cite{IntroBiology}).  Therefore,  calibrating  complicated ODE systems is of great interest and importance to both theorists and practitioners.

Owing to the superior performance of deep learning in modeling complicated data, deep neural networks have been actively used to reproduce dynamical systems (\cite{IntroDS1}). Deep neural networks, such as residual network (\cite{IntroResNet}) and discrete normalizing flows (\cite{IntroNormalingFlows}), can be considered as discrete dynamical systems. Recently, \cite{IntroNipsChen} propose a new family of continuous neural networks that extend the traditional discrete sequence of hidden layers to continuous-depth by using an ODE to parameterize the hidden units. \cite{IntroDS2} and \cite{IntroDS3} consider an autoencoder-based architecture to understand and predict  complex dynamical systems.

Despite advances in deep learning, most of the existing methods lack interpretability and their theoretical underpinnings are not well grounded. 
In this paper, we attempt to fill the gap and provide statistical justification for calibrating a complex system that can be characterized by multi-dimensional nonlinear ODEs. In particular, we are interested in scenarios where data, collected from continuous-time nonlinear ODEs,  are asynchronized,  irregularly spaced and are contaminated by measurement errors.

To start with, consider the following  multi-dimensional $\nu$th order ODE system in its general form:
\begin{equation}\label{higherordereq}
\frac{d^\nu}{dt^\nu}\bm{x}(t)=\bbf_0(\bm{x}(t),\bm{x}^{(1)}(t), \cdots, \bm{x}^{(\nu-1)}(t)), \quad 0\le t\le 1,%
\end{equation}
where $\bm{x}(t) = (x_1(t),\ldots,x_d(t))^{\top}\in\bbR^d$ and $\bm{x}^{(j)}(t)$ represents the $j$th derivative of $\bm{x}(t)$.    With $r_0 = \nu d$, $\bbf_0(\cdot) = (f_{0,1}(\cdot),\ldots,f_{0,d}(\cdot))^{\top}: \bbR^{r_0}\to \bbR^d$ is a $d$-dimensional function of $\bm{x}(t),\bm{x}^{(1)}(t), \cdots, \bm{x}^{(\nu-1)}(t)$, and it represents the unknown ground-truth, characterizing the interactions among $\bm{x}(t)$, $\bm{x}^{(1)}(t)$, $\cdots$, $\bm{x}^{(\nu-1)}(t)$.
When $\nu=1$, (\ref{higherordereq}) degenerates to a first order ODE system:
\begin{equation}\label{eq:1.1}
\frac{d}{dt}\bm{x}(t) = \bbf_0(\bm{x}(t)),  \quad 0\le t\le 1.
\end{equation}

Suppose that the $d$-dimensional continuous-time process $\bm{x}(t)$ is observed at discrete and possibly asynchronous time points; that is, $x_j(t)$ is observed at $0\le t_{j1} < t_{j2} < \ldots < t_{jn_j}\leq 1$, $j=1, \ldots, d$.  It is common that the collected data are contaminated by measurement error. Therefore, we use $y_{ji}$ to denote the data point we observe at time $t_{ji}$ and it relates to $x_j(t_{ji})$ in the following way:
\begin{equation}\label{eq:1.2}
y_{ji} = x_j(t_{ji})+\epsilon_{ji}, \quad i = 1, \ldots, n_j, ~ j = 1, \ldots, d,
\end{equation}
where $\epsilon_{ji} \sim N(0, \sigma_j^2)$  represents the measurement error at time $t_{ji}$.  The time stamps are allowed to be irregularly spaced and asynchronous, thereby, yielding different sample sizes $n_j$ for different components $j$.  The statistical task is to estimate the multi-dimensional nonlinear function $\bbf_0$ in \eqref{higherordereq} using the observed noisy data $\{\bm{y}_1, \ldots, \bm{y}_d\}$, where $\bm{y}_j = (y_{j1}, \ldots,y_{jn_j})^\top$. It is worthwhile to mention that our goal is to estimate $\bbf_0$ as a function of $\bm{x}(t),\bm{x}^{(1)}(t), \cdots, \bm{x}^{(\nu-1)}(t)$, not a function of $t$.

In general, it is computationally cumbersome to estimate $\bbf_0$ via conventional nonparametric techniques.  
To avoid this, we utilize deep neural networks and propose a two-stage nonparametric estimation procedure. In Stage 1, we use kernel approach to filter out the noise in $\{\bm{y}_1, \ldots, \bm{y}_d\}$ and obtain consistent estimators of $\bm{x}(t)$ and its high order derivatives $\bm{x}^{(j)}(t)$, denoted by $\widehat{\bm{x}}(t)$ and $\widehat{\bm{x}}^{(j)}(t)$, $j=1, 2, \ldots, \nu$, respectively (see Section \ref{sec.kernel}).  In Stage 2, we adopt a ReLU feedforward neural network to approximate $\widehat{\bm{x}}^{(\nu)}(t)$. By assuming that the function $\bbf_0(\cdot)$ enjoys a general modular structure with each modular component involving only a few input variables, we establish the consistency of the rectified linear unit (ReLU) feedforward neural network estimator  and derive its  convergence rate. In particular, we show that the rate is not subject to the dimension of the ODE system, but depends solely on the length and width of the neural network $\hat{\bbf}(\cdot)$ and the smoothness of the function $\bbf_0(\cdot)$ (See Section \ref{sec.relu}). As pointed out by \cite{hartford2017deep}, DNN is more computationally tractable than conventional nonparametric methods such as spline methods in the scenarios with high-dimensional features.

In light of (\ref{higherordereq}), each component of $\bm{x}(t)$ satisfies
\begin{equation}\label{higherordereq_single}
\frac{d^\nu}{dt^\nu}x_j(t)=f_{0,j}(\bm{x}(t),\bm{x}^{(1)}(t), \cdots, \bm{x}^{(\nu-1)}(t)), \quad 0\le t\le 1,
\end{equation}
for $j=1, \ldots, d$. Oftentimes the inputs, $\bm{x}(t),\bm{x}^{(1)}(t), \cdots, \bm{x}^{(\nu-1)}(t)$, do not contribute equally to $x^{(\nu)}_j(t)$. In some applications, the governing function $f_{0,j}$ is sparsely represented; in other words, only a small subset of the inputs $(x_1(t),\ldots,x_d(t), x^{(1)}_1(t),\ldots,x^{(1)}_d(t), \ldots, x^{(\nu-1)}_1(t),\ldots,x^{(\nu-1)}_d(t))$ is associated with $x^{(\nu)}_j(t)$. For example, biologists might have interest in recovering gene regulatory networks from noisy expression data where the cell regulation is only associated with a small set of genes. In such cases, it is critical to learn the governing equations as well as the associated coordinate system simultaneously.
Based on the interactions between feature (or variable) selection search and the learning model, traditional feature selection methods fall into three broad categories: filter methods, wrapper methods, and embedded methods. However, successful applications of these classical methods in the ODE setting are limited unless the ODE system is linear or parametric, or contains only lower order derivatives. Recently, \cite{lassonet} introduces a new neural network framework called LassoNet, which can capture the nonlinearity in $f_{0,j}$ nonparametrically with global feature selection. Motivated by the superior  performance of LassoNet, we present a accurate and computational feasibility method that is capable of identifying relevant input variables in (\ref{higherordereq_single}) after estimating $\bm{x}(t)$ and its derivatives from noisy data.

The rest of the paper is organized as follows. In Section \ref{Related_work}, we review some related work on governing equation estimation and variable selection in ODE systems, and point out where our work stands in the literature. Section \ref{Methodology} details the proposed method as well as its theoretical properties. In Section \ref{Simulation}, we conduct simulation study to assess the theoretical findings.  An application to the COVID  cases is presented in Section \ref{real data}.  Section \ref{Dissussion} concludes the paper. All the proofs are collected in the supplementary material.

{\bf Terminologies and Notation:} In this paper all vectors are column vectors. For two positive sequences $\{a_n\}$ and $\{b_n\}$, we say $a_n \lesssim b_n$ if there exists a positive constant $c$ such that $a_n \leq cb_n$ for all $n$, and $a_n \asymp b_n$ if $c^{-1}a_n \leq b_n \leq ca_n$  for some constant $c > 1$ and a sufficiently large $n$. For $\bm{x}=(x_1, \ldots, x_d)^{\top}$, let  $\|\bm{x}\|_2^2 = \bm{x}^{\top}\bm{x}$, $|\bm{x}| = (|x_1|, \ldots, |x_d|)$, $|\bm{x}|_\infty=\max_{i=1,\ldots,d}|x_i|$, and $|\bm{x}|_0=\sum_{i=1}^d \mathbbm{1}(x_i\neq0)$, where $\mathbbm{1}(\cdot)$ is the indicator function. For two $d$-dimensional vectors $\bm{x}=(x_1, \ldots, x_d)^{\top}$ and $\bm{y}=(y_1, \ldots, y_d)^{\top}$, we say $\bm{x} \lesssim \bm{y}$ if $x_i \lesssim y_i$, for $i=1,\ldots,d$.  For an $n\times n$ matrix $A = (a_{i,j})_{n\times n}$, let $\|A\|_\infty = \max_{i,j=1, \ldots, n}|a_{ij}|$ be the max norm of $A$ and  $\|A\|_0$ be the number of non-zero entries of $A$.
Let $\|f\|_2^2 = \int f(x)^2dx$ the $L_2$ norm of a real-valued function $f$ and $\||\bbf|_\infty\|_\infty$ the sup-norm for a $d$-dimensional function $\bbf$. We use $\floor{x}$ to represent the largest number less than $x$ and $\ceil{x}$ the smallest number greater than $x$, and use $a\wedge b$ and $a\vee b$ to represent the minimum and maximum of two numbers $a$ and $b$, respectively.


\section{Related Work} \label{Related_work}
In a parametric/semi-parametric setting where $\bbf_0(\cdot)$ is parameterized, the process of estimating the unknown parameters in $\bbf_0$ is the so-called inverse problem and has been widely studied in  statistical literature. For example, consider the first order ODE system in (\ref{eq:1.1}) and suppose that it involves an unknown parameter $\theta$, i.e.,$\frac{d\bm{x}(t;\theta)}{dt} = \bbf_0(\bm{x}(t;\theta);\theta)$. It has been discussed in, among other, \cite{IntroLSE} and \cite{IntroLSE1986} that $\theta$ can be consistently estimated by the least square estimator provided that the data are collected regularly and synchronously over time, i.e., $n_1 = n_2 =\ldots=n_d = n$ and $t_{1i} = t_{2i} = \ldots = t_{di}$ for $i=1,\dots,n$:
\begin{align*}
\hat{\theta}_{LSE} = \argmin_\theta \sum_{i=1}^n\sum_{j=1}^d (y_{ji}-x_j(t_{ji}))^2.  
\end{align*}
If the measurement errors $\epsilon_{ji}$ are normally distributed, then $\hat{\theta}_{LSE} $ coincides with the maximum likelihood estimator and is $\sqrt{n}$-consistent. Unfortunately, in most cases, $\hat{\theta}_{LSE}$ has no closed-form expression and tends to be computationally expensive. To overcome this issue, many other methods have been developed; see \cite{IntroWu2008}, \cite{PeterHall2014}, \cite{IntroBayes}, \cite{IntroWu2018}, and \cite{IntroBiometrika2020}, to name a few. However, they all suffer from the curse of dimensionality and can only deal with lower order derivatives.

In  cases where $\bbf_0$ cannot be summarized by a few low-dimensional parameters, calibrating $\bbf_0$ becomes more demanding.  Existing solutions attempt to impose extra assumptions in order to simplify the structure of $\bbf_0$. For instance, \cite{IntroNonparametric2} and \cite{IntroNonparametric1} assume an additive structure on $\bbf_0$, while \cite{IntroNonparametric3} considers $\bbf_0$ to be positive. To well preserve the structure of $\bbf_0$  so as to align with what is observed in practice,  we suggest a two-stage deep learning based method that estimates $\bbf_0$ nonparametrically by imposing a general modular structure on  $\bbf_0$ with each modular component involving only a few input variables. This method can handle higher order derivatives and recover the ODE system without being subject to the curse of dimensionality.

Statistical literature on variable selection in the context of ODE system largely focuses on linear, low dimensional, or lower order derivative ODE. For example, \cite{Featureselection1} only consider the first order derivative and estimate the governing function $\bbf_0$ as a linear combination of basis functions. In the work of \cite{IntroNonparametric1}, the authors also assume the derivative is only first order and the right-hand side of (\ref{higherordereq_single}) is additive, that is $f_{0,j} = \theta_{j,0} + \sum_{k=1}^{d}f_{0,j,k}(x_k(t))$ for some $\theta_{0,j} \in \bbR$. Recently, \cite{IntroWu2018} propose a new parameter estimation and variable selection method based on similarity transformation and separable least squares for large-scale systems; however, their approach can only be used in homogeneous linear system. Identifying relevant features for multi-dimensional complex ODE remains an open challenge.

Recently, deep neural network has been successfully applied in many fields, such as computer vision \citep{IntroResNet}, natural language processing \citep{bahdanau2014neural}, survival analysis \citep{li2022variable}, bioinformatics \citep{min2017deep}, recommendation systems \citep{zhang2019deep}, spatial statistics \citep{li2023semiparametric}, and variable selection \citep{lassonet, li2023deep}. Investigating theoretical properties of deep neural network is also of great interest to many statisticians and data scientists alike. In the statistical literature, \cite{schmidt-hieber} proves that using sparsely connected deep neural networks with ReLU activation function can achieve the minimax rate of convergence in a nonparametric setting. 
\cite{IntroLiang2020} obtains  convergence rates similar to \cite{schmidt-hieber} under  different regularity conditions. Similarly, \cite{IntroBauer} shows that multilayer feed-forward neural networks are able to circumvent the curse of dimensionality if the ground-truth $\bbf_0$ satisfies a generalized hierarchical interaction model.  A theoretical understanding of deep learning can also be achieved through approximation theory; for example, see \cite{IntroApproximation5}, \cite{IntroApproximation3}, \cite{IntroApproximation4}, \cite{IntroApproximation1}, \cite{IntroApproximation2}. Unlike the traditional function approximation theory that uses the aggregation of simple functions to approximate complicated ones, deep neural networks use the compositions of simple functions, which motivates us to assume that the function $\bbf_0$ satisfies a compositional structure.

\section{Methodology and Main Theorem} \label{Methodology}
In this section, we shall detail the two-stage estimation procedure and provide its theoretical underpinnings.  We also present a variant of the Stage-2 estimator to achieve variable selection in the context of ODE systems.

\subsection{Stage 1: Kernel Estimator}\label{sec.kernel}
Our goal in the first stage is to estimate $\bm{x}(t) = (x_1(t),\ldots,x_d(t))^{\top}$ and its $\nu$th derivative $\bm{x}^{(\nu)}(t), \nu\geq 1$, for $0\le t\le 1$, on the basis of the noisy observations $\{\bm{y}_1, \ldots, \bm{y}_d\}$. This is achieved via   kernel estimation  by casting \eqref{eq:1.2} as a nonparametric regression.

The classical kernel estimator of $x_j(t)$, $j=1,\ldots,d$, is given by $$\tilde{x}_j(t) = \sum_{i=1}^{n_j}\frac{t_{ji}-t_{j(i-1)}}{h_j}K(\frac{t-t_{ji}}{h_j}) \cdot y_j(t_{ji}),$$ where $t_{j0}=0$ and $h_j$ is a sequence of positive bandwidths satisfying $h_j\rightarrow 0$  and $n_jh_j\rightarrow\infty$ as $n_j \rightarrow \infty$, and   $K(\cdot)$ is a non-negative Lipschitz continuous kernel function satisfying
\begin{equation}\label{eqn.kernel1}
\int_{-\infty}^{\infty} K(x)dx = 1,  ~ \int_{-\infty}^{\infty} (K(x))^2 dx < \infty.
\end{equation}
Despite that $\tilde{x}_j(t) $ is consistent for $x_j(t)$ for $t\in (0,1)$ (see \cite{priestley1972non}), the classical kernel estimator does not lead to a consistent estimator of  $x_j^{(\nu)}(t)$. Therefore, we adopt the  boundary kernel function introduced in {\cite{Muller1984}} here.

We first estimate $x_j(t), j=1,\ldots,d, $ by
\begin{equation}\label{eq:2.1}
\widehat{x}_j(t) = \frac{1}{h_j}\sum_{i=1}^{n_j}\int_{s_{i-1}}^{s_i}K(\frac{t-u}{h_j})du \cdot y_j(t_{ji}),
\end{equation}
where  $0=s_0\leq s_1,\ldots, \leq s_{n_j}=1$, $s_i \in [t_{ji}, t_{j(i+1)}],$  $i=1,\ldots,n_j-1$, $h_j$ is a sequence of positive bandwidths satisfying $h_j\rightarrow 0$ and $n_jh_j\rightarrow\infty$ as $n_j \rightarrow \infty$. Moreover,  $K(\cdot)$ is a $\nu$ times differentiable kernel function that has a compact support on $[-\tau, \tau]$ with $K(-\tau) = K(\tau)=0$ and fulfills \eqref{eqn.kernel1}, and its $\nu$th derivative, $K_{\nu}$, meets the following requirements:
\begin{enumerate}[(i)]
    \item The support of $K_{\nu}$ is $[-\tau, \tau]$ and $\int_{-\tau}^{\tau} K_{\nu}(x) dx = 1$;
    \item  For constants $\beta\in \mathbb{R}$ and  $k\geq\nu+2$,
    \begin{equation}\label{eqn.nuder}
    \begin{aligned}
    \int_{-\tau}^{\tau} K_{\nu}(x)x^jdx=
            \left\{
             \begin{array}{ll}
             0   &j=\{0, \ldots, k-1\}\setminus \nu,\\
             (-1)^\nu \nu!,&j=\nu ,\\
             \beta, & j=k .
             \end{array}
            \right.
    \end{aligned}
    \end{equation}
\end{enumerate}
In order to obtain a consistent estimator of $\bm{x}^{(\nu)}(t)$ for $t\in [0,1]$, i.e., to eliminate boundary effects that become prominent when estimating derivatives, we introduce the modified kernel  $K_{\nu, q}$ with support $[-\tau, q\tau]$, for some  $q\in[0, 1]$, satisfying $K_{\nu, q} \rightarrow K_{\nu}$ as $q\rightarrow 1$.   Moreover, $K_{\nu, q}(x) $ satisfies \eqref{eqn.nuder} with a uniformly bounded $k$-th moment for $q$, and the asymptotic variance of $K_{\nu, q}$ is also bounded uniformly for $q$ (\cite{Muller1984}). Therefore, for $\kappa=1, \ldots, \nu$, the derivative ${x}^{(\kappa)}_j(t)$ is estimated by
\begin{equation}\label{eq:2.2}
\widehat{x}_j^{(\kappa)}(t) = \frac{1}{h_j^{\kappa+1}}\sum_{i=1}^{n_j}\int_{s_{i-1}}^{s_i}K_{\kappa, q}(\frac{t-u}{h_j})du \cdot y_j(t_{ji}), \quad  j = 1, 2, \ldots, d.
\end{equation}
\cite{Muller1984} and \cite{Gasser1983} have discussed the existence of such kernel functions $K(\cdot)$ and $K_{\nu, q}(\cdot) $, and have shown that if the sequence $\{s_i\}$ satisfies $\max_{i=1, \ldots, n_j}|s_i-s_{i-1}-n_j^{-1}|=O(n_j^{-\delta})$ for some $\delta>1$, then $\int_0^1\|\widehat{\bm{x}}^{(\nu)}(t) - \bbf_{0}(\bm{x}(t), \bm{x}^{(1)}(t), \ldots, \bm{x}^{(\nu-1)}(t))\|_2^2dt = O_P(n^{-2(k-\nu)/(2k+1)})$, where $n = \min\{n_1, \ldots, n_d\}$.

\subsection{Stage 2: Deep Neural Network Estimator}\label{sec.relu}
In Stage 2, we  use  a multilayer feedforward neural network to approximate the unknown ground-truth $\bbf_0(\cdot)$. We start off by introducing the definitions of H\" older smoothness and  compositional functions.

\begin{definition}\label{holder smoothness}
A function $g: \mathbb{R}^{r_0} \to \mathbb{R}$ is said to be $(\beta, C)$-H\" older smooth for some positive constants $\beta$ and $C$, if for every $\bm{\gamma}=(\gamma_1, \ldots, \gamma_{r_0})^{\top} \in \mathbb{N}^{r_0}$ the following two conditions hold: letting $\kappa = \sum_{i=1}^{r_0}\gamma_{i}$,
\begin{eqnarray}
\sup_{\bm{z}\in
\mathbb{R}^{r_0}}\bigg|\frac{\partial^{\kappa}g}{\partial z_1^{\gamma_1}\ldots \partial z_{r_0}^{\gamma_{r_0}}}(\bm{z})\bigg|\leq C,
\quad     \textrm{ if }\kappa\leq \floor{\beta},\nonumber
\end{eqnarray}
and
\begin{align*}
\bigg|\frac{\partial^{\kappa}g}{\partial
z_1^{\gamma_1}\ldots \partial
z_{r_0}^{\gamma_{r_0}}}(\bm{z})-\frac{\partial^\kappa g}{\partial
z_1^{\gamma_1}\ldots \partial
z_{r_0}^{\gamma_{r_0}}}(\widetilde{\bm{z}})\bigg| &\leq C\|\bm{z}-\widetilde{\bm{z}}\|_2^{\beta-\floor{\beta}}, \quad \textrm{if } \kappa=\floor{\beta},
\end{align*}
for $\bm{z},\widetilde{\bm{z}}\in \mathbb{R}^{r_0}$.
\end{definition}

For convenience, we say $g$ is $(\infty, C)$-H\" older smooth if $g$ is $(\beta, C)$-H\" older smooth for all $\beta>0$.  H\" older smoothness is commonly assumed for estimating regression functions nonparametrically in the literature (see, for instance, \cite{stone1985additive} and \cite{ferraty2006nonparametric}).

\begin{definition}\label{Compsitional stucture}
A function $f: \mathbb{R}^{r_0} \to \mathbb{R}$ is said to have a compositional structure with parameters $(L_*, \bm{r}, \bm{\tilde{r}}, \bm{\beta}, \bm{a}, \bm{b}, \bm{C})$ for $L_*\in \mathbb{Z}_+$, $\bm{r}=(r_0, \ldots, r_{L_*+1})^\top$ $\in \mathbb{Z}_+^{L_*+2}$ with $r_{L_*+1}=1$, $\bm{\tilde{r}}=(\tilde{r}_0,\ldots, \tilde{r}_{L_*})^\top \in \mathbb{Z}_+^{L_*+1}$, $\bm{\beta}=(\beta_0,\ldots, \beta_{L_*})^\top \in \mathbb{R}_+^{L_*+1}$, $\bm{a}=(a_0,\ldots, a_{L_*+1})^\top$, $\bm{b}=(b_0,\ldots, b_{L_*+1})^\top \in \mathbb{R}^{L_*+2}$, and $\bm{C}=(C_0,\ldots, C_{L_*})^\top \in \mathbb{R}_+^{L_*+1}$, if
\begin{equation*}
     f(\bm{z})=\bm{g}_{L_*}\circ\ldots\circ \bm{g}_1 \circ \bm{g}_0(\bm{z}),\quad\quad\quad   \bm{z} \in [a_0, b_0]^{r_0},
\end{equation*}
where for $i=0, 1, \ldots, L_*$, $\bm{g}_i=(g_{i,1},\ldots, g_{i,r_{i+1}})^\top: [a_i, b_i]^{r_i}\to [a_{i+1}, b_{i+1}]^{r_{i+1}}$ for some $|a_i|, |b_i|\leq C_i$, and the functions $g_{i,j}: [a_i, b_i]^{\tilde{r}_i} \to [a_{i+1}, b_{i+1}]$ are $(\beta_i, C_i)$-H\" older smooth only relying on $\tilde{r}_i$ variables, with $\tilde{r}_i \le r_i$.
\end{definition}

This definition is well connected with the structure of deep neural network (see Definition \ref{DNN}), where each composition can be viewed as a hidden layer in a neural network. In practice, people tend to use dropout to avoid overfitting such that only a few nodes are ``active''. Thus, it is natural to assume that each component of $\bm{g}_i$ only relies on $\tilde{r}_i$ variables. Without loss of generality, we can always assume $C_i>1$, $i=0, \ldots, L_*$. Denote by $\mathcal{CS}(L_*, \bm{r}, \bm{\tilde{r}}, \bm{\beta},\bm{a}, \bm{b}, \bm{C})$ the class of compositional functions defined above. By definition, any function in $\mathcal{CS}(L_*, \bm{r}, \bm{\tilde{r}}, \bm{\beta},\bm{a}, \bm{b}, \bm{C})$ is composed of $L_* + 1$ layers, and in the $i$-th layer, $i=0,\ldots,L_*$, there are only $\tilde{r}_i$ ``active" variables. This implicitly assumes a sparsity structure in each layer, which  avoids the curse of dimensionality. 
Functions in each layer $g_{i,j}: [a_i, b_i]^{\tilde{r}_i} \to [a_{i+1}, b_{i+1}]$ are $(\beta_i, C_i)$-H\" older smooth and it is not difficult to verify that the composed function $f$ is also H\" older smooth.
It is pointed out by, for instance, \cite{schmidt-hieber} that a compositional function can be approximated by a neural network with any order of accuracy.  Due to their popularity, compositional functions have been adopted by, among many others, \cite{IntroBauer},  \cite{schmidt-hieber}, \cite{DIVE}, \cite{Kohle2019}, \cite{Functional_shang},  and \cite{optdnn} to study nonparametric regression problems.

It is well known that the optimal convergence rate of a $(\beta, C)$-H\" older smooth regression function defined on $\mathbb{R}^p$ is $n^{-\frac{2\beta}{2\beta+p}}$. When $p$ is large, this rate suffers from the curse of dimensionality. However, for any $d$-dimensional function $\bm{f} = (f_1, \ldots, f_d)^{\top}$ with each component $f_j$ belonging to $\mathcal{CS}(L_*, \bm{r}, \bm{\tilde{r}}, \bm{\beta},\bm{a}, \bm{b}, \bm{C})$, the ``intrinsic'' dimension of $\bm{f}$ that actually determines the rate convergence should be less than $r_0$ owing to the sparsity of $\bm{f}$. Moreover, from the definition of $\mathcal{CS}(L_*, \bm{r}, \bm{\tilde{r}}, \bm{\beta},\bm{a}, \bm{b}, \bm{C})$, each component of $\bm{f}$, $f_j$, is made up of H\" older smooth functions with the level of smoothness varying from layer to layer, and we use the concept  the intrinsic smoothness to characterize the overall smoothness of $f_j$.   Next, we introduce the definitions of the intrinsic smoothness and intrinsic dimension of the compositional functions.

\begin{definition}\label{p*t*}
For $f \in \mathcal{CS}(L_*, \bm{r}, \bm{\tilde{r}}, \bm{\beta},\bm{a}, \bm{b}, \bm{C})$, the intrinsic smoothness and intrinsic dimension of $f$ are defined as:
\begin{eqnarray}
     \beta^*=\beta_{i^*}^*\quad \textrm{ and }\quad r^*=\tilde{r}_{i^*},\nonumber
\end{eqnarray}
respectively, where $\beta_i^*=\beta_i\prod_{s=i+1}^{L_*}(\beta_s\wedge 1)$ for $i=0,\ldots, L_*$, and $i^*=\argmin_{0\leq i \leq L_*}\beta_i^*/\tilde{r}_i$, with the convention $\prod_{s=L_*+1}^{L_*}(\beta_s\wedge 1)=1$.
\end{definition}

It is  worth noting that the order of H\" older smoothness of a function $f$ is less than its intrinsic smoothness.
Throughout the paper, we assume the ground-truth $f_{0,j} \in \mathcal{CS}(L_*, \bm{r}, \bm{\tilde{r}}, \bm{\beta},\bm{a}, \bm{b}, \bm{C})$, $j=1, \ldots,d$. This assumption is not restrictive, as the compositional structure covers a wide collection of functions. Here are some examples.

\begin{example} \label{Example1}
(Homogeneous Linear ODE system) Consider the following dynamical system
\begin{align*}
    \frac{d^{\nu}}{dt^{\nu}}
        \begin{bmatrix}
            x_1(t) \\
            \vdots \\
            x_d(t)
        \end{bmatrix}
    =
        \begin{bmatrix}
            a_{1,1} & \ldots & a_{1,\nu d}\\
            \vdots & \ddots & \vdots\\
            a_{d,1} & \ldots & a_{d,\nu d}
        \end{bmatrix}
        \begin{bmatrix}
            x_1(t) \\
            \vdots \\
            x^{(\nu-1)}_d(t)
        \end{bmatrix}.
\end{align*}
Each component $x^{(\nu)}_j$ has a compositional structure with $L_*=0$, $\bm{r} = (\nu d, 1)^{\top}$, $\bm{\tilde{r}}=\nu d$, and $\bm{\beta} = \infty$. Therefore, $\beta^*=\infty$ and $r^*=\nu d$.
\end{example}

The next example presents a more general case than Example \ref{Example1}.
\begin{example}
(Additive Model) Define $x^{(\nu)}_i = g_i(\sum_{j=1}^{d}\sum_{k=0}^{\nu-1}f_{j,k}(x_j^{(k)}))$, where $g_i(\cdot)$ is $(\beta_g, C_g)$-H\" older smooth and $f_{j,k}(\cdot)$ is $(\beta_f, C_f)$-H\" older smooth. By definition, $x^{(\nu)}_i$ can be written as a composition of three functions $x^{(\nu)}_i=h_2\circ h_1\circ h_0$, with $h_0(x_1, \ldots, x_d^{(\nu-1)}) = (f_{1,0}(x_1),\ldots,f_{d,\nu-1}(x_d^{(\nu-1)}))$, $h_1(x_1, \ldots, x_{\nu d}) =\sum_{i=1}^{\nu d}x_i$, and $h_2(x) = g_i(x)$. Here $L_*=2, \bm{r}=(\nu d, \nu d, 1, 1), \bm{\tilde{r}} = (1, \nu d, 1), \bm{\beta} = (\beta_h, \infty, \beta_g), \beta^* = \min(\beta_h, \beta_g)$, and $r^*=1$.
\end{example}

The intuition behind Definition \ref{Compsitional stucture} comes from the structure of feedforward neural network.  We next introduce the ReLU feedforward neural network that is widely used in the deep learning literature.  For $\bm{v} = (v_1,\ldots, v_r)^{\top} \in \bbR^r$, define the shifted activation function $\sigma_{\bm{v}}(\bm{x})= (\sigma(x_1-v_1),\ldots,\sigma(x_r-v_r))^{\top}$,  where $\sigma(s) := \max\{0,s\}$ and $\bm{x} = (x_1,\ldots,x_r)^{\top}  \in \bbR^r$.

\begin{definition}\label{DNN}
A ReLU feedforward neural network $\bbf(\bm{x}; W, v)$ is defined as
\begin{equation}
\bbf(\bm{x}; W, v):=W_{L}\sigma_{\bm{v}_{L}} \ldots W_1 \sigma_{\bm{v}_1} W_0 \bm{x}, \quad \bm{x}\in\bbR^{p_0},\label{fnn}
\end{equation}
where $W_{l}\in \mathbb{R}^{p_{l+1}\times p_{l}}, 0\le l\le L,$ are the weight matrices, $\bm{v}_{l}\in \mathbb{R}^{p_l}$, $1\le l\le L$, are referred to as biases, and $p_0, p_1, \ldots, p_{L+1}$ are positive integers.
\end{definition}
The ReLU feedforward neural network is parameterized by $(W_j)_{j=0,\ldots,L}$ and $(\bm{v}_j)_{j=1,\ldots,L}$, where $L$ determines the number of hidden layers. The width vector $\bm{p}=(p_0, \ldots, p_{L+1})$ specifies the number of units in each layer, i.e., the width of the network.  Clearly, a feedforward neural network belongs to $\mathcal{CS}(L_*, \bm{r}, \bm{\tilde{r}}, \bm{\beta},\bm{a}, \bm{b}, \bm{C})$, with the $i$-th hidden layer  viewed as $\bm{g}_i$.

In this paper, we consider two subclasses of ReLU feedforward neural networks:
\begin{align*}
\mathcal{F}_1(L, \bm{p})&:=  \{\bbf(\bm{x}; W, v)  \textrm{ of form } \eqref{fnn}: \max_{j=0,\ldots,L} \|W_j\|_\infty + |\bm{v}_j|_\infty \leq 1\} \numberthis \label{class1}
\end{align*}
where $\bm{v}_0$ is a vector of zeros, and
\begin{align*}
\mathcal{F}_2(L, \bm{p}, \tau, F) &:= \{\bbf(\bm{x}; W, v) \in \mathcal{F}_1(L, \bm{p}): & \sum_{j=0}^{L}(\|W_j\|_0 + |\bm{v}_j|_0) \leq \tau, \||\bbf|_\infty\|_\infty \leq F\}. \numberthis  \label{class2}
\end{align*}
The subclass (\ref{class1}) comprises the fully connected networks with bounded parameters, and it is not empty in that we can always rescale the weights by dividing all the weights and the biases by their maximum. In practice, people tend to use dropout as a regularization technique to prevent overfitting, i.e., randomly setting parts of neurons to zero. So it is reasonable to consider a sparse neural network as specified in (\ref{class2}). The two subclasses of neural networks are also considered in \cite{schmidt-hieber} and \cite{Functional_shang} which use  i.i.d. data and functional data, respectively, to  estimate regression function nonparametrically. In practical applications, achieving precise control over the exact number of inactive nodes in a neural network can be challenging. To address this, we adopt an alternative approach by introducing an $L_1$ penalty during the optimization process. This penalty effectively regulates the number of active nodes within each layer, allowing us to attain a desirable level of sparsity in the network. This strategy provides a more practical and flexible means of controlling network complexity and achieving optimal performance. Similar approaches are used in the literature, such as those presented in \cite{ma2019transformed, Functional_shang, lassonet}.

Next, we present the estimator of the unknown groun-truth $\bbf_{0}(\cdot)$ in (\ref{higherordereq}) via a sparsely connected deep neural network. Let $\widehat{\bm{x}}(t) = (\widehat{x}_1(t),\ldots,\widehat{x}_d(t))^{\top}$ and $\widehat{\bm{x}}^{(\nu)}(t)  = (\widehat{x}_1^{(\nu)}(t),\ldots,\widehat{x}_d^{(\nu)}(t))^{\top}$ the kernel estimators of $\bm{x}(t)$ and $\bm{x}^{(\nu)}(t)$ obtained from Stage 1, where $\widehat{x}_j(t)$ and $\widehat{x}^{(\nu)}_j(t)$ are respectively given in (\ref{eq:2.1}) and (\ref{eq:2.2}). The idea is to search for a member in $ \mathcal{F}_2(L, \bm{p}, \tau, F)$ that well approximates $\bbf_0(\bm{x}(t), \bm{x}^{(1)}(t), \cdots, \bm{x}^{(\nu-1)}(t))$. Specifically, the ``best" estimator of $\bbf_0(\bm{x}(t), \cdots, \bm{x}^{(\nu-1)}(t))$, or equivalently $\bm{x}^{(\nu)}(t)$, is obtained by minimizing
\begin{equation}\label{fnn.est}
\int_{0}^{1}    \left\|\widehat{\bm{x}}^{(\nu)}(t)-\bbf(\widehat{\bm{x}}(t),\widehat{\bm{x}}^{(1)}(t), \ldots, \widehat{\bm{x}}^{(\nu-1)}(t); W, v)\right\|_2^2 dt
\end{equation}
over all $\bbf(\bm{z}; W, v)  \in \mathcal{F}_2(L, \bm{p}, \tau, F)$ with $\bm{p} = (r_0, p_1, \ldots, p_{L}, d)$ and $\bm{z}\in\bbR^{r_0}$. The resulting estimator is denoted by $\widehat{\bbf}(\bm{z}; \widehat{W}, \widehat{v})$. The entire two-stage estimation procedure is summarized in Algorithm \ref{algo1}.
Its validity is justified by the following theorem that establishes the consistency of $\widehat{\bbf}(\bm{z}; \widehat{W}, \widehat{v})$ as an estimator of $ \bbf_{0}(\bm{x}(t),  \ldots, \bm{x}^{(\nu-1)}(t))$ and its convergence rate.

\begin{algorithm}
\caption{Training $\widehat{\bbf}(\bm{z}; \widehat{W}, \widehat{v})$}
\label{algo1}
\begin{algorithmic}
\State \textbf{Input}:  observed values $\{\bm{y}_1, \ldots, \bm{y}_d\}$, feed-forward neural network $\bbf(\bm{z}; W, v)  \in \mathcal{F}_2(L, \bm{p}, \tau, F)$, number of epochs $E$,  and learning rate $\alpha$
\For{each component $j$, $j=1, \ldots, d$, }
\State Estimate $x_j(t)$ and its higher order derivative $x_j^{(\kappa)}(t), \kappa = 1,\ldots,\nu,$ through (\ref{eq:2.1}) and \eqref{eq:2.2}, respectively:
$\widehat{x}_j(t) = \frac{1}{h_j}\sum_{i=1}^{n_j}\int_{s_{i-1}}^{s_i}K(\frac{t-u}{h_j})du \cdot y_j(t_{ji})$ and
$\widehat{x}_j^{(\kappa)}(t) = \frac{1}{h_j^{\kappa+1}}\sum_{i=1}^{n_j}\int_{s_{i-1}}^{s_i}K_{\kappa, q}(\frac{t-u}{h_j})du \cdot y_j(t_{ji})$
\State where the tuning parameter can be determined by cross validation
\EndFor
\State Use $(\widehat{\bm{x}}(t),\widehat{\bm{x}}^{(1)}(t), \ldots, \widehat{\bm{x}}^{(\nu-1)}(t))$ as the input to fit $\widehat{\bm{x}}^{(\nu)}(t)$ through minimizing the loss function defined as $\ell(W, v) = \int_{0}^{1}    \left\|\widehat{\bm{x}}^{(\nu)}(t)-\bbf(\widehat{\bm{x}}(t),\widehat{\bm{x}}^{(1)}(t), \ldots, \widehat{\bm{x}}^{(\nu-1)}(t); W, v)\right\|_2^2 dt$
\State Initialize $W, v$
\For{$e \in \{1, \ldots, E\}$}
\State Compute  gradient of the loss $\ell(W, v)$ with respect to the parameters
\State $(W, v), \nabla _W \ell(W, v), \nabla _v \ell(W, v)$ using back-propagation
\State Update $W \leftarrow W - \alpha\nabla _W \ell(W, v), v\leftarrow v - \alpha\nabla _v \ell(W, v)$
\EndFor
\State \textbf{Output}: $\widehat{\bbf}(\bm{z}; \widehat{W}, \widehat{v})$
\end{algorithmic}
\end{algorithm}

\begin{theorem}\label{thm.nn}
Suppose the  true functions $f_{0,j} \in \mathcal{CS}(L_*, \bm{r}, \bm{\tilde{r}}, \bm{\beta},\bm{a}, \bm{b}, C)$ with the intrinsic smoothness and intrinsic dimension $\beta^*$ and $r^*$, respectively, $j=1, \ldots,d$. Consider the subclass $\mathcal{F}_2(L, \bm{p}, \tau, F)$.  Let $\eta = \max_{i=0,\ldots, L_*}(r_{i+1}(\tilde{r}_i + \ceil{\beta_i}))$ and $N =\min_{i=1,\ldots,L+1} p_i$.  Assume that there exist some constants $\tilde{C}_i$, $i=0,\ldots,L^*$, only depending on $\bm{C}, \bm{a}$, and $\bm{b}$ such that $N \geq 6\eta\max_{i=0,\ldots,L_*}(\beta_i+1)^{\tilde{r}_i} \vee (\tilde{C}_i+1)e^{\tilde{r}_i}$, $\tau\lesssim LN$, and $F \geq \max_{i=0,\ldots,L^*}(C_i, 1)$.  Then we have
\[
\int_{0}^{1}\|\widehat{\bbf}(\bm{x}(t), \ldots, \bm{x}^{(\nu-1)}(t); \widehat{W}, \widehat{v})
      - \bbf_{0}(\bm{x}(t),   \ldots, \bm{x}^{(\nu-1)}(t))\|_2^2dt = O_P(\varsigma_n),
\]
\[
\int_{0}^{1}\|\widehat{\bbf}(\widehat{\bm{x}}(t),   \ldots, \widehat{\bm{x}}^{(\nu-1)}(t);  \widehat{W}, \widehat{v}) - \bbf_{0}(\bm{x}(t),  \ldots, \bm{x}^{(\nu-1)}(t))\|_2^2 dt = O_P(\varsigma_n),
\]
where $\varsigma _{n}= (1+N^L)n^{-2(k-\nu)/(2k+1)} + (N2^{-L})^{2\prod_{l=1}^{L_*}\beta_l\wedge1} + N^{-\frac{2\beta^*}{r^*}}.$
\end{theorem}

To simplify the notation, we have assumed in Theorem \ref{thm.nn} that the intrinsic smoothness and intrinsic dimension of the components of $\bbf_{0}$ are all the same.  Theorem \ref{thm.nn} is still valid if this assumption is relaxed, by setting $\beta^*$ and $r^*$ to the lower bound of the intrinsic smoothness and the upper bound of the intrinsic dimension, respectively, over the components of $\bbf_{0}$.

The convergence rate in Theorem \ref{thm.nn} contains three parts: the first part $n^{-2(k-\nu)/(2k+1)}$ is the approximation error due to the kernel method; the second part $N^Ln^{-2(k-\nu)/(2k+1)}$ corresponds to perturbation error of neural network; the last part $(N2^{-L})^{2\prod_{l=1}^{L_*}\beta_l\wedge1} + N^{-\frac{2\beta^*}{r^*}}$  is associated with the approximation  of $\mathcal{F}_2(L, \bm{p}, \tau, F)$ to $\bm{f}_0$. Note that $\varsigma _{n}$ is free of the input dimension $d$ which helps us get around the curse of dimensionality.
The consistency of $\widehat{\bbf}(\bm{z}; \widehat{W}, \widehat{v})$ is guaranteed, if the tuning parameters are suitably selected, for instance, $L\asymp (\log n)^{1/2}$ and $N\asymp e^{(\log{n})^{1/4}}$, since the latter implies $\varsigma _{n} \rightarrow 0$ as $n\rightarrow\infty$.

\subsection{Variable Selection} \label{sec_VS}

Owing to the sparseness assumption in $\mathcal{CS}(L_*, \bm{r}, \bm{\tilde{r}}, \bm{\beta},\bm{a}, \bm{b}, \bm{C})$, it is natural to assume $f_{0,j}$ is also sparsely represented. In many cases, the right-hand side of (\ref{higherordereq}) or \eqref{higherordereq_single} involves only a few input variables, rather than the entire set $(x_1(t),\ldots,x_d(t)$, $x^{(1)}_1(t),\ldots,x^{(1)}_d(t)$, $\ldots$, $x^{(\nu-1)}_1(t),\ldots,x^{(\nu-1)}_d(t))$.  Moreover, the set of relevant input variables may well vary from component to component; that is,  the active input variables that relate to $x^{(\nu)}_j(t)$ are not necessarily the same as those to $x^{(\nu)}_{j'}(t)$, $j'\neq j$.  However, the estimator $\widehat{\bbf}(\bm{z}; \widehat{W}, \widehat{v})$ obtained by minimizing (\ref{fnn.est}) cannot select the relevant subset from the collection of the input variables.
\cite{lassonet} propose a new feature/variable selection framework for neural networks by adding a penalized input-to-output residual layer and selecting the active features only if the corresponding weights are nonzero.
To facilitate variable selection in \eqref{higherordereq_single}, inspired by \cite{lassonet}, we modify the ReLU feedforward neural network considered in Stage 2 and introduce a new class of neutral network $f_{\textrm{vs}}(\bm{z}; \bm{\theta}, W, v)$ for each component, which is defined as
\begin{align}\label{f_vs}
f_{\textrm{vs}}(\bm{z}; \bm{\theta}, W, v) = \bm{\theta}^{\top}\bm{z} + f(\bm{z};W, v)
\end{align}
where $f(\bm{z}; W, v)  \in \mathcal{F}_2(L, \bm{p}, \tau, F)$ with $\bm{p} = (r_0, p_1, \ldots, p_L, 1)$ and $\bm{\theta}, \bm{z}\in\bbR^{r_0}$. The difference between $f_{\textrm{vs}}(\bm{z}; \bm{\theta}, W, v)$ and the standard feed-forward neural network is the inclusion of the residual layer $\bm{\theta}^{\top}\bm{z}$ which makes it considerably easier in tackling the vanishing gradient problem.

Assume that $f_{0,j}$ involves only a subset of the input variables $(x_1(t),\ldots,x_d(t)$, $x^{(1)}_1(t),\ldots,x^{(1)}_d(t)$, $\ldots$, $x^{(\nu-1)}_1(t),\ldots,x^{(\nu-1)}_d(t))$.  We first estimate $x_1(t),\ldots,x_d(t)$ and their derivatives via the kernel method outlined in Section \ref{sec.kernel}, and then solve the following optimization problem in order to obtain  the estimator of $f_{0,j}$: $\widehat{f}_{\textrm{vs},j}(\bm{z}; \widehat{\bm{\theta}}, \widehat{W}, \widehat{v}) = \widehat{\bm{\theta}}^{\top}\bm{z} + f(\bm{z};\widehat{W}, \widehat{v})$, with
\begin{align}\label{fnn.variable.selection}
(\widehat{\bm{\theta}}, \widehat{W}, \widehat{v}) = &\argmin \int_{0}^{1}    \left\| \widehat{x}_j^{(\nu)}(t) - f_{\textrm{vs}}(\widehat{\bm{x}}(t),\widehat{\bm{x}}^{(1)}(t), \ldots, \widehat{\bm{x}}^{(\nu-1)}(t); \bm{\theta}, W, v)\right\|_2^2 dt + \lambda\|\bm{\theta}\|_1, \notag\\
&\textrm{ subject to } \|W_{0,i}\|_\infty \leq M|\theta_i|, i = 1, \ldots, r_0,
\end{align}
In \eqref{fnn.variable.selection}, $\theta_i$ is the $i$-component of $\bm{\theta}\in\bbR^{r_0}$ and $W_{0,i}$ contains the weights  for the $i$-th input variable in the first hidden layer. There are two tuning parameters in the objective function: $\lambda$ and $M$, which penalize the linear and nonlinear components simultaneously. The constraint $\|W_{0,i}\|_\infty \leq M|\theta_i|$ plays a vital role, in that $M$ leverages the  effect of the $i$-th input variable, thereby capturing the non-linearity in the data. When $M=0$, the neural network part in $f_{\textrm{vs}}(\bm{z}; \bm{\theta}, W, v)$ vanishes and $f_{\textrm{vs}}(\bm{z}; \bm{\theta}, W, v) = \bm{\theta}^{\top}\bm{z}$, and thus \eqref{fnn.variable.selection} degenerates to standard LASSO. When $M \rightarrow \infty$, we get a feed-forward network with an $\ell_1$ penalty on the residual layer.  Following the optimization procedure outlined in \cite{lassonet}, we optimize the objective function \eqref{fnn.variable.selection} using hierarchical proximal gradient descent. Hierarchical proximal operator is a widely used algorithm for solving composite optimization problems where the objective function can be decomposed into a sum of several components, each with its own convex structure. For further insights, refer to \cite{chambolle2011first, bauschke10convex}. It is worth mentioning that without domain knowledge, it is difficult to determine the value of $M$.  In practical implementation, we  use cross validation to select $M$, a common practice in machine learning algorithms. The detailed pseudocode is summarized in Algorithm \ref{algo2}.

\begin{algorithm}
\caption{Training $\widehat{f}_{\textrm{vs},j}(\bm{z}; \widehat{\bm{\theta}}, \widehat{W}, \widehat{v})$}
\label{algo2}
\begin{algorithmic}
\State \textbf{Input}: observed values $\{\bm{y}_1, \ldots, \bm{y}_d\}$, $f(\bm{z}; W, v)  \in \mathcal{F}_2(L, \bm{p}, \tau, F)$, number of epochs $E$,  learning rate $\alpha$, hyper-parameter $M$, and path multiplier $\epsilon$
\For{each component $j$, $j=1, \ldots, d$, }
\State Estimate $x_j(t)$ and its higher order derivative $x_j^{(\kappa)}(t), \kappa = 1,\ldots,\nu$ through (\ref{eq:2.1}) and \eqref{eq:2.2}, respectively:
$\widehat{x}_j(t) = \frac{1}{h_j}\sum_{i=1}^{n_j}\int_{s_{i-1}}^{s_i}K(\frac{t-u}{h_j})du \cdot y_j(t_{ji})$ and
$\widehat{x}_j^{(\kappa)}(t) = \frac{1}{h_j^{\kappa+1}}\sum_{i=1}^{n_j}\int_{s_{i-1}}^{s_i}K_{\kappa, q}(\frac{t-u}{h_j})du \cdot y_j(t_{ji})$
\State where the tuning parameter can be determined by cross validation
\EndFor
\State Use $(\widehat{\bm{x}}(t),\widehat{\bm{x}}^{(1)}(t), \ldots, \widehat{\bm{x}}^{(\nu-1)}(t))$ as the input to fit $\widehat{x}^{(\nu)}_j(t)$ through minimizing the loss function defined as $\ell = \ell(\bm{\theta}, W, v) = \int_{0}^{1}    \left\|\widehat{x}_j^{(\nu)}(t)-f_{\textrm{vs},j}(\widehat{\bm{x}}(t),\widehat{\bm{x}}^{(1)}(t), \ldots, \widehat{\bm{x}}^{(\nu-1)}(t); \bm{\theta}, W, v)\right\|_2^2 dt + \lambda\|\bm{\theta}\|_1, \notag$
\State Initialize $\bm{\theta}, W, v$, $\lambda=\epsilon, k=r_0$
\While{$k > 0$}
\State Update $\lambda \leftarrow (1+\epsilon)\lambda$
\For{$e \in \{1, \ldots, E\}$}
\State Compute gradients $\nabla _{\bm{\theta}} \ell, \nabla _W \ell, \nabla _v \ell$ using back-propagation
\State Update $\bm{\theta} \leftarrow \bm{\theta} - \alpha\nabla _{\bm{\theta}}\ell, W \leftarrow W - \alpha\nabla _W \ell, v \leftarrow v - \alpha\nabla _v\ell$
\State Update $(\bm{\theta}, W_0) \leftarrow \textrm{Hier-Prox}(\theta, W_0, \alpha\lambda, M)$, where Hier-Prox is provided in Algorithm \ref{algo3} in the Appendix.
\EndFor
\State Update $k$ to be the number of non-zero elements of $\bm{\theta}$
\EndWhile
\State \textbf{Output}: $\widehat{f}_{\textrm{vs},j}(\bm{z}; \widehat{\bm{\theta}}, \widehat{W}, \widehat{v})$
\end{algorithmic}
\end{algorithm}

\subsection{Computational Complexity}
In this section, we assess the computational cost of the proposed procedure. We break down the time complexity into two main components: the kernel estimator in Stage 1 and the deep neural network estimator in Stage 2. It is noteworthy that both stages encompass integral computations and can leverage various numerical algorithms with a time complexity of $O(n)$, such as Simpson's method and the Newton-Cotes method. For a comprehensive review of these numerical algorithms, refer to \cite{burden2015numerical}. For the fist stage, it is evident that the time complexity for each time-point, denoted as $\widehat{x}_j(t)$ and $\widehat{x}_j^{(\kappa)}(t)$ where $j = 1, \ldots, d$ and $\kappa = 1, \ldots, \nu$, amounts to $O(n_j^2)$. Let $\tilde{n} = \max\{n_1, \ldots, n_d\}$. In terms of training $\widehat{\bbf}(\bm{z}; \widehat{W}, \widehat{v})$ in second stage, the time complexity per training sample hinges on the number of layers $L$, the neuron count in each layer represented by $\bm{p}=(p_0, \ldots, p_{L+1})$, and the number of epochs $E$. Both the feed-forward and backpropagation phases involve matrix multiplications. Given that the weight matrix $W_{l}$ in the $l$th layer has dimension ${p_{l+1}\times p_{l}}$, the time complexity within each layer is $O(p_{l+1}p_{l})$. Since the overall structure comprises $L$ layers, the time complexity can be approximated as $O(E\sum_{l=1}^Lp_{l+1}p_{l})$. The training $\widehat{f}_{\textrm{vs},j}(\bm{z}; \widehat{\bm{\theta}}, \widehat{W}, \widehat{v})$ involves the hierarchical proximal gradient descent. Notably, since only the first layer's weight matrix is penalized, and in accordance with \cite{bauschke10convex, lassonet}, the computational complexity of hierarchical proximal gradient descent is controlled by $O(p_0p_1\log(p_0p_1))$, which is negligible in comparison to the computation required for updating other parameters. Thus, the computational complexity of training $\widehat{f}_{\textrm{vs},j}(\bm{z}; \widehat{\bm{\theta}}, \widehat{W}, \widehat{v})$ is the same as training $\widehat{\bbf}(\bm{z}; \widehat{W}, \widehat{v})$.

\section{Numerical Experiments}\label{Simulation}

In this section, we carry out  simulation studies to assess the finite sample performance of the proposed two-stage estimation procedure and variable selection algorithm for two different designs of  ODE systems. In both designs, we use the same network architecture with length $L=3$ and width $N=30$. The tuning parameters in the first stage of estimation procedure are selected by cross validation. 
The dropout probability is set to $0.2$ to avoid overfitting. Summary statistics from each simulation setting are calculated based on 100 independent simulation runs.

\subsection{Simulation Study 1} \label{simulations1}

In the first simulation study (i.e., Design 1), we consider a second order nonlinear ODE system:
\begin{align*}
x_1^{(2)}(t) =& 2\frac{x_1(t) }{x_3(t) } + 4x_4^{(1)}(t) - x_3(t) x_4(t), \\
x_2^{(2)}(t)  =& -x_4^{(1)}(t),  \quad x_3^{(2)}(t)  = 2,  \\
x_4^{(2)}(t) =& x_2^{(1)}(t),  \quad x_5^{(2)}(t) = x_5(t), \\
x_6^{(2)}(t)  =& -x_5^2(t) x_6(t) +x_6^{(1)}(t), \\
x_7^{(2)}(t) =& x_7^{(1)}(t)x_2^{(1)}(t) - x_2(t) x_7(t), \\
x_8^{(2)}(t) =& -(x_8^{(1)}(t))^2.
\end{align*}
The noisy data are computed via  $y_{ji} = x_j(t_{i})+\epsilon_{ji},$  for $i = 1, \ldots, n$ and $j = 1, \ldots, 8$, with $t_i = i/n$ and $\epsilon_{ji}$ sampled independently from $N(0,\sigma^2)$.  We consider three noise levels: $\sigma = 0.2, 0.5, 0.8$, and the sample size $n$ is chosen from $n \in \{100, 200, 500\}$. We treat the first 80\% of the data, i.e., $\{y_{ji}, i=1, 2, \ldots, 0.8n\}$ as the training set, and the remaining are put aside as the test set.

In this experiment,  we aim to evaluate the performance of the procedure described in Sections \ref{sec.kernel} and \ref{sec.relu}. To this end, the following  metrics are employed:
\begin{align*}
M_1 =& \int_{0.8}^{1} \left\|\widehat{\bbf}(\bm{x}(t), \bm{x}^{(1)}(t), \ldots, \bm{x}^{(\nu-1)}(t)) - \bbf_{0}(\bm{x}(t),   \ldots, \bm{x}^{(\nu-1)}(t)) \right\|_2dt, \\
M_2 =& \int_{0.8}^{1} \left\|\widehat{\bbf}(\widehat{\bm{x}}(t), \widehat{\bm{x}}^{(1)}(t), \ldots, \widehat{\bm{x}}^{(\nu-1)}(t)) - \bbf_{0}(\bm{x}(t),   \ldots, \bm{x}^{(\nu-1)}(t)) \right \|_2 dt, \\
M_3 =& \max\limits_{j=1,\ldots,d}\int_{0.8}^{1} \left| \widehat{f}_j(\bm{x}(t), \bm{x}^{(1)}(t), \ldots, \bm{x}^{(\nu-1)}(t))
      - f_{0,j}(\bm{x}(t),   \ldots, \bm{x}^{(\nu-1)}(t)) \right| dt, \\
M_4 =& \max\limits_{j=1,\ldots,d}\int_{0.8}^{1} \left|\widehat{f}_j(\widehat{\bm{x}}(t), \widehat{\bm{x}}^{(1)}(t), \ldots, \widehat{\bm{x}}^{(\nu-1)}(t)) - f_{0,j}(\bm{x}(t),   \ldots, \bm{x}^{(\nu-1)}(t)) \right| dt.
\end{align*}
Metrics $M_1$ and $M_3$ measure the differences between the target and the second-stage deep neural network estimator in the $L_2$ norm and in the max norm, respectively, while $M_2$ and $M_4$ gauge the overall performance of the proposed two-stage estimator via these two norms.  All the metrics are computed using the test data.


Table \ref{TableSim1} reports the averaged values of the four metrics and their standard deviations (in parentheses) over 100 replications for different combinations of noise level and sample size.  As expected, the deviations measured by $M_2$ ($M_4$) are greater than those by $M_1$ ($M_3$) unanimously.  Our results also show that the accuracy of the estimator diminishes the noise-to-signal ratio increases.  Nevertheless,  a larger sample size always yields a smaller deviation across all cases, which is in line with the theoretical results.

\subsection{Simulation Study 2} \label{simulations2}

In the second simulation (Design 2), 
we suppose that the true data generating process is governed by the following homogeneous first order linear ODE system:
\[ \bm{x}^{(1)}(t) = A\bm{x}(t) + \bm{b},  \quad 0\le t\le 1,\]
where $A$ is a $d\times d$ sparse matrix and $\bm{b}$ is a $d$-dimensional vector referred to as the initial value. Elements of $\bm{b}$ are sampled from the uniform distribution $U(0,1)$. We further define a set $J = \{j_1, j_2, j_3, j_4, j_5\}$, where $j_k, k=1,\ldots,5$, are randomly sampled from $\{1, \ldots, d\}$ without replacement. The coefficient matrix $A = (a_{ij})_{d\times d}$ is sparse and its elements are defined in the following way: for $i=1, \ldots, d$,  if $j \in J$, then $a_{ij}$ is sampled from $U(0,1)$; otherwise, $a_{ij} = 0$.  In other words, each row of $A$ has five non-zero values that are sampled from the uniform distribution. It is easy to verify that for $j=1, \ldots,d$,  $f_{0,j} \in \mathcal{CS}(L_*, \bm{r}, \bm{\tilde{r}}, \bm{\beta},\bm{a}, \bm{b}, C)$ with $L_*=0$, $\bm{r} = (d, 1)^{\top}$, $\bm{\tilde{r}}=5$, $\bm{\beta} = \infty, \beta^*=\infty$, and $r^*=5$. For simplicity, we set $n_1 = n_2 = \ldots = n_d = n$. The ``observed" data are generated from $y_{ji} = x_j(t_{i})+\epsilon_{ji},$  $i = 1, \ldots, n, j = 1, \ldots, d$, where $t_i = i/n$ and $\epsilon_{ji}$ are sampled from $N(0,1)$.  The sample size $n$ and dimension $d$ are chosen to be $n = 100, 200, 500$ and $d=10, 100, 1000$. The train-test split is the same as the first simulation.

The objective of this study is twofold. Firstly,  we want to check how the dimension $d$ impacts the accuracy of the two-stage estimation procedure for the linear system, and to evaluate the variable selection algorithm. Secondly, we would like to compare our method with GRADE by \cite{IntroNonparametric1} which utilizes a nonparametric approach to estimate high-dimensional additive ordinary differential equations and achieve variable selection.

For each $j=1,\ldots,d$, we use the method described in Section \ref{sec_VS} to find $\widehat{f}_{\textrm{vs},j}(\bm{z})$. Let $\widehat{\bbf} = (\widehat{f}_{\textrm{vs},1}, \ldots, \widehat{f}_{\textrm{vs},d})^\top$.  Similar to Section \ref{simulations1}, we use $M_1, M_2, M_3$, and $M_4$ to assess the estimation errors. In order to evaluate the performance of variable selection, we consider the following two metrics: 
\begin{itemize}
\item MinSize = the minimum number of selected variables to includes all true variables
\item $\textrm{ProbAll}$ = the success rate that the selected five variables are all true variables
\end{itemize}
These two metrics are widely used in feature selection literature (see,  for instance, \cite{zhong2015iterative} and \cite{RunzeLi}). By definition, MinSize is expected to be at least five; and the closer to five, the better the procedure. The metric $\textrm{ProbAll}$ measures the sensitivity (true positive) rate of the method in detecting the true variables. A higher value of $\textrm{ProbAll}$ is desirable, as this indicates  there is a higher chance that the algorithm will pick all the true variables.

The simulation results that summarize the estimation errors are reported in Table \ref{TableSim2}, while the variable selection results are  in Table \ref{TableSim3_variable_selection}. Table \ref{TableSim2} indicates that (i) although increasing dimension exerts an adverse impact on the accuracy of the estimator, a large sample size can easily offset the impact;
(ii) our method performs better than GRADE across all the values of  $n$ and $d$ under consideration.  As for the variable selection accuracy, we observe from Table \ref{TableSim3_variable_selection} that our method yields a much smaller minimum selection size than GRADE, albeit  greater than five, and that the probability of selecting all the true variables is higher when using our method than that of GRADE.  The simulation results in both tables demonstrate the effectiveness of our method, especially when  dimension $d$ is large.

\begin{table}
\caption{Simulation results of Design 1 for the proposed method. The table reports the averaged values of the four metrics and their standard deviations (in parentheses) over 100 replications for different combinations of noise level $\sigma$ and sample size $n$.
}
\label{TableSim1}
\centering
\begin{tabular}{llccc}
\hline
&           & $\sigma =  0.2 $  & $\sigma =  0.5$   & $\sigma =  0.8$    \\ \cline{2-5}
& $n$ = 100 & 0.050 (0.007) & 0.081 (0.008) & 0.142 (0.017) \\
Metric $M_1$ & $n$ = 200 & 0.044 (0.005) & 0.051 (0.006) & 0.085 (0.011)  \\
& $n$ = 500 & 0.034 (0.003) & 0.041 (0.004) & 0.049 (0.005)
\end{tabular}%

\begin{tabular}{llccc}
\hline
& $n$ = 100 & 0.061 (0.011) & 0.122 (0.016) & 0.188 (0.021) \\
Metric $M_2$ & $n$ = 200 & 0.052 (0.006) &0.092 (0.009) & 0.126 (0.014) \\
& $n$ = 500 & 0.049 (0.004) & 0.079 (0.006) & 0.097 (0.008)
\end{tabular}%

\centering
\begin{tabular}{llccc}
\hline
& $n$ = 100 & 0.224 (0.024) & 0.274 (0.034) & 0.476 (0.051) \\
Metric $M_3$ & $n$ = 200 & 0.196 (0.017) & 0.211 (0.023) & 0.285 (0.029) \\
& $n$ = 500 & 0.181 (0.012) & 0.201 (0.015) & 0.212 (0.018)
\end{tabular}%

\begin{tabular}{llccc}
\hline
& $n$ = 100 & 0.226 (0.037) & 0.375 (0.041) & 0.614 (0.071) \\
Metric $M_4$ & $n$ = 200 & 0.204 (0.028) & 0.302 (0.033) & 0.410 (0.047) \\
& $n$ = 500 & 0.197 (0.017) & 0.255 (0.021) & 0.302 (0.028) \\ \hline
\end{tabular}%
\end{table}


\begin{table}
\caption{Simulation results of Design 2  for the proposed method and GRADE. The table reports the averaged values of the four metrics and their standard deviations (in parentheses) over 100 replications for different combinations of dimension $d$ and sample size $n$.}
\label{TableSim2}
\centering
\scalebox{0.75}{
\begin{tabular}{lllccc}
\hline
&          & Methods    &  $d =  10 $       & $d =  100$   & $d=1000$ \\\cline{2-6}
&$n$ = 100 & Our Method & 0.331 (0.034) & 0.387 (0.042)  & 0.756 (0.081)\\
&          & GRADE      & 0.448 (0.053) & 0.554 (0.062)  & 1.144 (0.131)\\\cline{2-6}

Metric $M_1$ & $n$ = 200 & Our Method & 0.198 (0.019) & 0.206 (0.022) &  0.241 (0.021)\\
&          & GRADE      & 0.309 (0.031) & 0.352 (0.047)&  0.383 (0.055)\\\cline{2-6}

&$n$ = 500 & Our Method & 0.103 (0.005) & 0.111 (0.005) & 0.126 (0.009)\\
&          & GRADE      & 0.169 (0.010) & 0.182 (0.017) & 0.199 (0.025) \\
\end{tabular}%
}
\scalebox{0.75}{
\begin{tabular}{lllccc}
\hline
&$n$ = 100 & Our Method & 0.498 (0.052)& 0.559 (0.076)& 1.004 (0.144)\\
&          & GRADE      & 0.732 (0.087)& 0.876 (0.101)& 1.423 (0.199)\\\cline{2-6}

Metric $M_2$&$n$ = 200 & Our Method & 0.239 (0.022)& 0.258 (0.031) & 0.301 (0.034)\\
&          & GRADE      & 0.357 (0.038)& 0.383 (0.067)& 0.418 (0.072)\\\cline{2-6}

&$n$ = 500 & Our Method & 0.134 (0.013)& 0.140 (0.011) & 0.177 (0.025)\\
&          & GRADE      & 0.202 (0.017)& 0.254 (0.031) & 0.309 (0.039)\\
\end{tabular}%
}
\centering
\scalebox{0.75}{
\begin{tabular}{lllccc}
\hline
&$n$ = 100 & Our Method & 1.01 (0.108) & 2.364 (0.184) & 8.406 (0.633)\\
&          & GRADE      & 1.763 (0.205)& 3.758 (0.445) & 12.214  (1.913)\\\cline{2-6}
Metric $M_3$&$n$ = 200 & Our Method & 0.682 (0.058)& 1.086
(0.103)& 2.008 (0.124)\\
&          & GRADE      & 1.148 (0.133)& 1.823 (0.243)& 3.456 (0.536)\\\cline{2-6}
&$n$ = 500 & Our Method & 0.343 (0.044)& 0.701 (0.088)& 1.019 (0.111)\\
&          & GRADE      & 0.507 (0.114)& 1.068 (0.148)& 1.516 (0.299)\\
\end{tabular}%
}

\scalebox{0.75}{
\begin{tabular}{lllccc}
\hline
&$n$ = 100 & Our Method & 1.348 (0.138)& 2.920 (0.380) &9.150 (0.510)\\
&          & GRADE      & 1.997 (0.227)& 4.010 (0.494) &14.317 (2.366)\\\cline{2-6}
Metric $M_4$&$n$ = 200 & Our Method & 0.821 (0.122)& 1.271 (0.149)& 2.273 (0.273)\\
&          & GRADE      & 1.225 (0.159)& 1.994 (0.273)& 3.933 (0.630)\\\cline{2-6}
&$n$ = 500 & Our Method & 0.394 (0.032) & 0.759 (0.053)& 1.109 (0.099)\\
&          & GRADE      & 0.639 (0.133) & 1.123 (0.172)& 1.727 (0.334)\\
\hline
\end{tabular}%
}
\end{table}


\begin{table}
\caption{Simulation results of Design 2  for the proposed method and GRADE. The table reports the averaged values of the two metrics that assess the variable selection accuracy and their standard deviations (in parentheses) over 100 replications for different combinations of dimension $d$ and sample size $n$.
}
\label{TableSim3_variable_selection}
\centering
\scalebox{1}{
\begin{tabular}{llllll}
\hline
&& Methods                      & $d$ =  10       & $d$ = 100       & $d$ = 1000   \\
\hline
&$n$ = 100 & Our Method         & 5.78 (1.06) & 6.37 (1.17)&  7.26 (1.26)\\
&          & GRADE              & 6.99 (1.23) & 8.09 (1.32)&  9.49 (1.48)\\\cline{2-6}
MinSize &$n$ = 200 & Our Method & 5.46 (0.94) & 5.98 (1.13)&  6.84 (1.21)\\
&          & GRADE              & 6.73 (1.19) & 7.59 (1.28)&  9.11 (1.43)\\\cline{2-6}
&$n$ = 500 & Our Method         & 5.32 (0.89) & 5.69 (1.09)&  6.36 (1.16)\\
&          & GRADE              & 6.19 (1.02) & 7.27 (1.22)&  8.38 (1.38)\\
\hline
\end{tabular}%
}
\centering
\scalebox{1}{
\begin{tabular}{llllll}
&& Methods & $d$ =  10       & $d$ = 100    & $d$ = 1000   \\
\hline
&$n$ = 100 & Our Method           & 0.92  & 0.82  & 0.67  \\
&          & GRADE                & 0.83  & 0.71  & 0.49  \\\cline{2-6}
ProbAll &$n$ = 200 & Our Method   & 0.94  & 0.86  & 0.72  \\
&          & GRADE                & 0.85  & 0.75  & 0.53  \\\cline{2-6}
&$n$ = 500 & Our Method           & 0.96  & 0.92  & 0.79  \\
&          & GRADE                & 0.89  & 0.81  & 0.55  \\
\hline
\end{tabular}%
}
\end{table}

\section{Real Data Analysis} \label{real data}

\begin{figure*}[h]
\includegraphics[scale=.32]{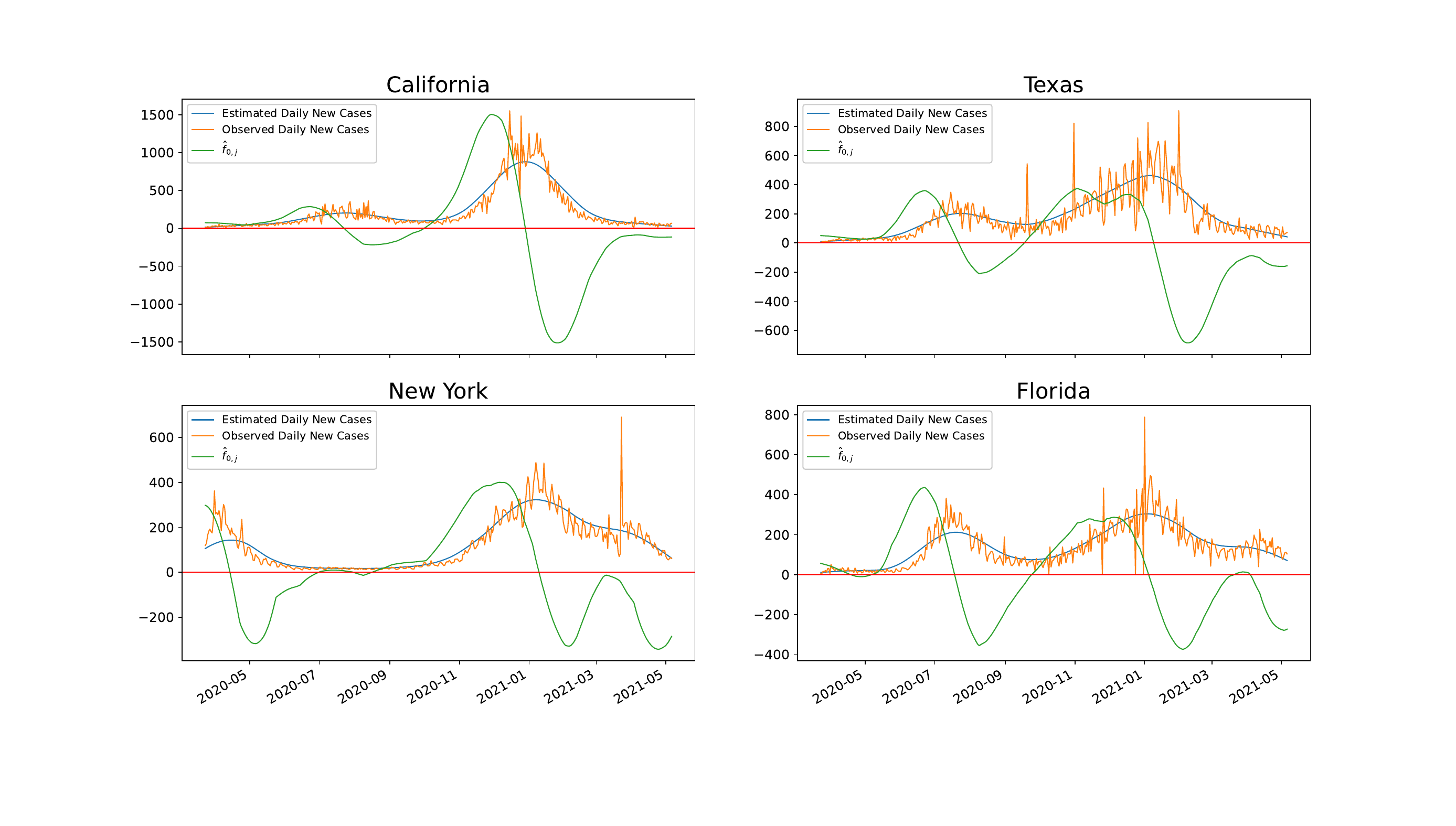} 
\caption{The orange curve depicts the observed daily new COVID cases from 03/23/2020 to 05/07/2021 in  California, Texas, New York State, and Florida, respectively.  The  blue curve is the estimated daily new cases, while the green one is the estimated growth rates.}
\label{Four States}
\end{figure*}

In this section, we illustrate our proposed method by the COVID-19 infection cases. The data are downloaded from \emph{New York Times. (2021)} \url{https://github.com/nytimes/covid-19-data}. This dataset consists of daily new  COVID-19 cases reported in individual states across the United States from 03/23/2020 to 05/07/2021, with a total of $411$ observations for each state. As an example, the orange curves in the four panels of Figure \ref{Four States} depict the number of the recorded daily new cases  in  California, Texas, New York, and Florida, respectively.

Here, we would like to use a system of ODEs to characterize the rate of changes of new cases over the 50 states {\it simultaneously}. Specifically, we  consider a 50-dimensional second order ODE system, as in  \eqref{higherordereq} with $\bm{x}(t)=(x_1(t),\ldots,x_{50}(t))^{\top}$ representing the number of COVID-19  cases at time $t$.  With the daily new cases, we first calculate the  daily cases accumulated since 03/23/2020 for each state, i.e.,  $y_{ji}$ in \eqref{eq:1.2} for $i=1, 2, \ldots, 411$, $j=1, 2, \ldots, 50$.  We then  use the proposed estimation approach to estimate $\bm{x}(t)$, its derivatives, and the function $\bbf_0$. Specifically, we use a neural network with length $L=3$ and width $N=50$ with dropout probability equals to $0.2$. The filtered daily new cases $\widehat{\bm{x}}(t)$ are shown as blue curves in Figure \ref{Four States} for the four selected states, while the deep neural network estimator $\widehat\bbf_0$ is shown in green. The results demonstrate the effectiveness and foreseeability of our method: when the estimated growth rate reaches the highest value, the daily new cases will  peak roughly one month later.

\begin{figure*}[h]
\includegraphics[scale=.3]{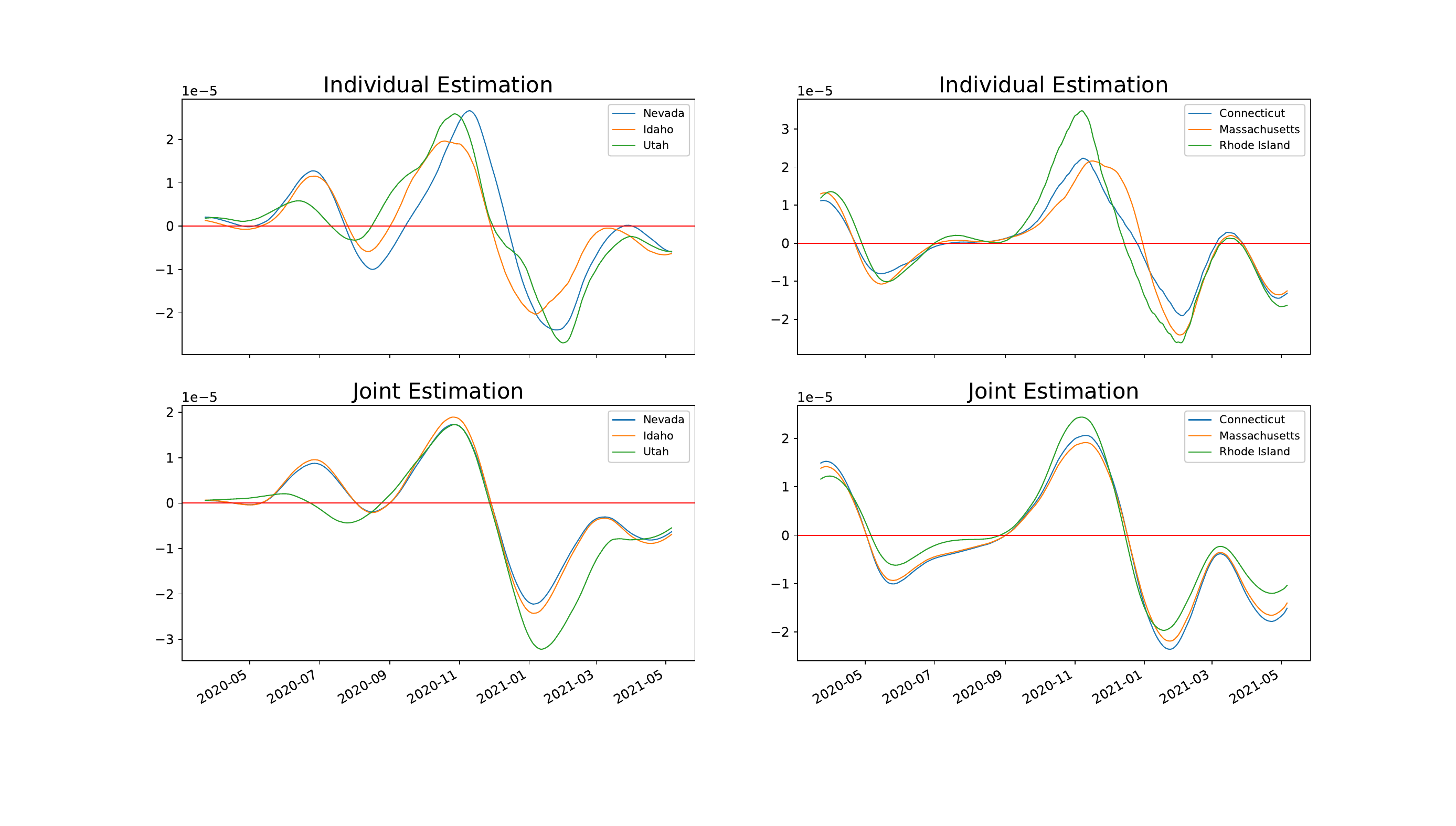}
\caption{A comparison between individual estimation (top two panels) and joint estimation (bottom two panels). Left two panels: western states. Right two panels: northeastern states.}
\label{Comparison}
\end{figure*}

To further illustrate the strength of  our approach, we compare  the proposed method with the traditional nonparametric estimation of the derivative of the regression function that estimates the growth rate of each state separately. The top two panels in Figure \ref{Comparison} show the  growth rates that are estimated separately for three states in the western  (Utah, Nevada, and Idaho) and northeastern (Connecticut, Rhode Island, and Massachusetts) U.S., respectively.  The  growth rates estimated using our method are depicted in the bottom two panels. To facilitate the comparison, we standardize the growth rate by dividing it by its respective population size.  It is apparent that there is a discrepancy between the two estimates, and our estimates are more consistent with what is expected:  geographically adjacent states should have strong interactions, thereby sharing similar growth rates. This further shows  the superiority of our method, in that it  effectively  incorporates the interactive processes among the neighboring states.

To better understand the interactions of these states, we apply our variable selection method described in Section \ref{sec_VS} to the data from  Massachusetts.  The top three states that are associated with Massachusetts are: New York State, Rhode Island, and Connecticut.

\section{Conclusion} \label{Dissussion}
In this study, we have introduced a novel nonparametric approach for estimating large-scale ordinary differential equation (ODE) systems from noisy data, particularly in cases involving intricate and nonlinear relationships among input variables. Our methodology addresses the challenge of accurately recovering ODE structures, even in the presence of complex underlying functions.

We have demonstrated the consistent capability of our proposed method to recover ODE structures, even when the true functional relationships are complex. The adoption of the state-of-the-art framework developed by \cite{lassonet} for neural network feature selection enhances our approach's applicability, allowing it to excel even in scenarios with nonlinear ODE systems. Rigorous simulation studies and a real data analysis have substantiated the validity and effectiveness of our methods.

However, our work is not without its limitations. One notable challenge lies in the substantial computational resources demanded by training deep neural networks, particularly in the context of high-dimensional systems. Finding strategies to balance computational efficiency with accuracy will be a crucial avenue for future research. Moreover, our current formulation assumes a specific noise distribution in the data. A more robust approach should be explored to accommodate a broader range of noise distributions, thereby enhancing the method's robustness and practicality.

Furthermore, while our approach excels in estimating complex functions, extracting meaningful biological, physical, or chemical insights from the estimated parameters remains a challenge. To enhance the interpretability of the obtained ODE structures, future research could focus on developing post-processing techniques that facilitate the extraction of valuable domain-specific knowledge from the estimated models.

Looking forward, exciting opportunities lie in the integration of our approach with causal inference techniques. This integration could shed light on the causal relationships inherent in ODE systems, paving the way for more nuanced and informed insights. Additionally, incorporating domain knowledge and prior information through appropriate regularization terms could enhance the accuracy and interpretability of the estimated ODE models.

In conclusion, our work contributes to advancing the field of ODE system estimation through a flexible and accurate methodology. While acknowledging its limitations, we have identified various directions for future research that hold promise in enhancing the methodology's robustness, efficiency, interpretability, and integration with other powerful analytical techniques. These avenues offer a rich landscape for researchers to explore and continue pushing the boundaries of accurate and interpretable ODE modeling.

\clearpage
  \bibliographystyle{chicago}      
  \bibliography{ref}

\newpage
\appendix

\section{Proof of Theorem 1}

The proof of Theorem 1 requires some preliminary lemmas.
\begin{lemma} \label{lemma 1}
For any $f: \mathbb{R}^{r_0} \to \mathbb{R} \in \mathcal{CS}(L_*, \bm{r}, \bm{\tilde{r}}, \bm{\beta},\bm{a}, \bm{b}, \bm{C})$, integer $m>0$ and $N \geq \max_{i=0,\ldots,L_*}(\beta_i+1)^{\tilde{r}_i} \vee (\tilde{C}_i+1)e^{\tilde{r}_i})$, there exists a neural network
\[
f_* \in \mathcal{F}_2(L, (r_0, 6\eta N,\ldots,6\eta N, 1), \sum_{i=0}^{L_*}r_{i+1}(\tau_i + 4), \infty),
\]
where
\begin{align*}
\tilde{C}_i &= \sum_{i=0}^{i}C_i\frac{b_i-a_i}{b_{i+1}-a_{i+1}}, i = 0,\ldots,L_*-1,\\
\tilde{C}_{L_*} &= \sum_{i=0}^{L_*}C_i\frac{b_i-a_i}{b_{i+1}-a_{i+1}}+b_{L_*} - a_{L_*}\\
L &= 3L_* + \sum_{i=0}^{L_*}L_i,\\
L_i &= 8 + (m+5)(1+\ceil{\log_2(\tilde{r}_i\vee\beta_i)}),\\
\tau_i &\leq 141(\tilde{r}_i + \beta_i + 1)^{3+\tilde{r}_i}N(m+6),\\
\eta &= \max_{i=0,\ldots, L_*}(r_{i+1}(\tilde{r}_i + \ceil{\beta_i})),
\end{align*}
 such that
\[
\| f_* - f\|_{\infty} \leq C_{L_*}\prod_{l=0}^{L_*-1}(2C_l)^{\beta_{l+1}}\sum_{i=0}^{L_*}\left((2\tilde{C}_i+1)(1+\tilde{r}^2_i + \beta_i^2)6^{\tilde{r}}N2^{-m} + \tilde{C}_i3^{\beta_i}N^{-\frac{\beta_i}{\tilde{r}_i}}\right)^{\prod_{l=i+1}^{L_*}\beta_l\wedge1}.
\]
\end{lemma}
\begin{proof}
Suppose
\begin{equation*}
     f(\bm{z})=\bm{g}_{L_*}\circ\ldots\circ \bm{g}_1 \circ \bm{g}_0(\bm{z}),\quad\quad \textrm{ for } \bm{z} \in [a_0, b_0]^{r_0}
\end{equation*}
where $\bm{g}_i=(g_{i,1},\ldots, g_{i,r_{i+1}})^\top: [a_i, b_i]^{r_i}\to [a_{i+1}, b_{i+1}]^{r_{i+1}}$ for some $|a_i|, |b_i|\leq C_i$ and the functions $g_{i,j}: [a_i, b_i]^{\tilde{r}_i} \to [a_{i+1}, b_{i+1}]$ are $(\beta_i, C_i)$-H\" older smooth. For $i=0,\ldots, L_*-1$, the domain and range of $\bm{g}_i$ are $[a_i, b_i]^{r_i}$ and $[a_{i+1}, b_{i+1}]^{r_{i+1}}$. In the first step, we will rewrite $f$ as the composition of functions $\bm{h}_i :=(h_{i,1},\ldots, h_{i,r_{i+1}})^\top$ whose domain and range are $[0, 1]^{r_i}$ and $[0, 1]^{r_{i+1}}$ by linear transformation. That is, we define
\begin{align*}
\bm{h}_i(\bm{z}) &:= \frac{\bm{g}_i((b_i - a_i)\bm{z} - a_{i+1})}{b_{i+1} - a_{i+1}},\quad\quad \textrm{ for } \bm{z} \in [0, 1]^{r_i}, i = 0,\ldots, L_*-1\\
\bm{h}_{L_*}(\bm{z}) &:= \bm{g}_{L_*}((b_{L_*} - a_{L_*})\bm{z}+a_{L_*}),\quad\quad \textrm{ for } \bm{z} \in [0, 1]^{r_{L_*}}.
\end{align*}
And the following equality holds
\begin{equation*}
     f(\bm{z})=\bm{g}_{L_*}\circ\ldots\circ \bm{g}_1 \circ \bm{g}_0(\bm{z})=\bm{h}_{L_*}\circ\ldots\circ \bm{h}_1 \circ \bm{h}_0(\frac{\bm{z}-a_0}{b_0 - a_0}),\quad\quad \textrm{ for } \bm{z} \in [a_0, b_0]^{r_0}
\end{equation*}

Since $g_{i,j}: [a_i, b_i]^{\tilde{r}_i} \to [a_{i+1}, b_{i+1}]$ are all  $(\beta_i, C_i)$-H\" older smooth, it follows that $h_{i,j}: [0, 1]^{\tilde{r}_i} \to [0, 1]$ are all  $(\beta_i, \tilde{C}_i)$-H\" older smooth, where $\tilde{C}_i$ is a constant only depends on $\bm{a}, \bm{b}, \bm{C}$, i.e., $\tilde{C}_i = \sum_{i=0}^{i}C_i\frac{b_i-a_i}{b_{i+1}-a_{i+1}}, i = 0,\ldots,L_*-1, \tilde{C}_{L_*} = \sum_{i=0}^{L_*}C_i\frac{b_i-a_i}{b_{i+1}-a_{i+1}}+b_{L_*} - a_{L_*}$

By Theorem 5 in \cite{schmidt-hieber}, for any integer $m\geq1$ and $N \geq \max_{i=0,\ldots,L_*}(\beta_i+1)^{\tilde{r}_i} \vee (\tilde{C}_i+1)e^{\tilde{r}_i}$, there exist a network
\[
\tilde{h}_{i,j} \in \mathcal{F}_2(L_i, (\tilde{r}_i, 6(\tilde{r}_i + \ceil{\beta_i})N,\ldots,6(\tilde{r}_i + \ceil{\beta_i})N,1), \tau_i, \infty),
\]
with $L_i = 8 + (m+5)(1+\ceil{\log_2(\tilde{r}_i\vee\beta_i)}), \tau_i\leq141(\tilde{r}_i + \beta_i + 1)^{3+\tilde{r}_i}N(m+6)$, such that
\[
\|\tilde{h}_{i,j} - h_{i,j}\|_\infty \leq (2\tilde{C}_i+1)(1+\tilde{r}^2_i + \beta_i^2)6^{\tilde{r}_i}N2^{-m} + \tilde{C}_i3^{\beta_i}N^{-\frac{\beta_i}{\tilde{r}_i}}.
\]

Note that the value of $\tilde{h}_{i,j}$ is $(-\infty, \infty)$, so we define $h^*_{i,j} := \sigma(-\sigma(-\tilde{h}_{i,j}+1)+1)$ by add two more layers $\sigma(1-x)$ to restrict $h^*_{i,j}$ into the interval $[0, 1]$. This introduces two more layers and four more parameters and the by the fact that $h_{i,j} \in [0, 1]$, we have $h^*_{i,j}\in \mathcal{F}_2(L_i+2, (\tilde{r}_i, 6(\tilde{r}_i + \ceil{\beta_i})N,\ldots,6(\tilde{r}_i + \ceil{\beta_i})N,1), \tau_i + 4, \infty)$ and
\[
\|h^*_{i,j} - h_{i,j}\|_\infty \leq \|\tilde{h}_{i,j} - h_{i,j}\|\leq (2\tilde{C}_i+1)(1+\tilde{r}^2_i + \beta_i^2)6^{\tilde{r}_i}N2^{-m} + \tilde{C}_i3^{\beta_i}N^{-\frac{\beta_i}{\tilde{r}_i}}.
\]

We further parallelize all $(h^*_{i,j})_{j=1,\ldots,r_{i+1}}$ together, then we get $\bm{h}^*_{i} := (h^*_{i, 1}, \ldots, h^*_{i, r_{i+1}})^\top \in \mathcal{F}_2(L_i+2, (r_i, 6r_{i+1}(\tilde{r}_i + \ceil{\beta_i})N,\ldots,6r_{i+1}(\tilde{r}_i + \ceil{\beta_i})N,r_{i+1}), r_{i+1}(\tau_i + 4), \infty)$. Moreover, we construct the composite network $f_* := \bm{h}^*_{L_*}\circ\ldots\circ\bm{h}^*_1\circ\bm{h}^*_0 \in \mathcal{F}_2(3L_* + \sum_{i=0}^{L_*}L_i, (r_0, 6\eta N,\ldots,6\eta N, 1), \sum_{i=0}^{L_*}r_{i+1}(\tau_i + 4), \infty)$, where $\eta = \max_{i=0,\ldots, L_*}(r_{i+1}(\tilde{r}_i + \ceil{\beta_i}))$.

By Lemma 3 in \cite{schmidt-hieber}, we conclude the following inequality holds
\begin{align*}
\|f - f_* \|_\infty &= \|\bm{h}_{L_*}\circ\ldots\circ \bm{h}_1 \circ \bm{h}_0 -  \bm{h}^*_{L_*}\circ\ldots\circ\bm{h}^*_1\circ\bm{h}^*_0\|_\infty\\
&\leq C_{L_*}\prod_{l=0}^{L_*-1}(2C_l)^{\beta_{l+1}}\sum_{i=0}^{L_*}\||\bm{h}_i - \bm{h}^*_i|_\infty\|_\infty^{\prod_{l=i+1}^{L_*}\beta_l\wedge1}\\
&\leq C_{L_*}\prod_{l=0}^{L_*-1}(2C_l)^{\beta_{l+1}}\sum_{i=0}^{L_*}\left((2\tilde{C}_i+1)(1+\tilde{r}^2_i + \beta_i^2)6^{\tilde{r}}N2^{-m} + \tilde{C}_i3^{\beta_i}N^{-\frac{\beta_i}{\tilde{r}_i}}\right)^{\prod_{l=i+1}^{L_*}\beta_l\wedge1}\\
&\leq C_{L_*}\prod_{l=0}^{L_*-1}(2C_l)^{\beta_{l+1}}\sum_{i=0}^{L_*}((2\tilde{C}_i+1)(1+\tilde{r}^2_i + \beta_i^2)6^{\tilde{r}}N2^{-m})^{\prod_{l=i+1}^{L_*}\beta_l\wedge1} + \\
 &\quad C_{L_*}\prod_{l=0}^{L_*-1}(2C_l)^{\beta_{l+1}}\sum_{i=0}^{L_*}(\tilde{C}_i3^{\beta_i}N^{-\frac{\beta_i}{\tilde{r}_i}})^{\prod_{l=i+1}^{L_*}\beta_l\wedge1}
\end{align*}
\end{proof}

\emph{\textbf{Proof of Theorem 1:}}\\
Without loss of generality, we only consider the case when $n$ is sufficiently large.
Let
\[
\hat{\bbf}(\bm{z},\hat{W}, \hat{v}) := \argmin_{\bbf \in \mathcal{F}_2(L, \bm{p}, \tau, F)}\int_0^1
    \left\|\widehat{\bm{x}}^{(\nu)}(t)-\bbf(\widehat{\bm{x}}(t),\widehat{\bm{x}}^{(1)}(t), \ldots, \widehat{\bm{x}}^{(\nu-1)}(t), W, v)\right\|_2^2dt, \bm{z}\in\bbR^{r_0},
\]

\[
    \tilde{\bbf}(\bm{z},\tilde{W}, \tilde{v}) := \argmin_{\bbf \in \mathcal{F}_2(L, \bm{p}, \tau, F)}\int_0^1
    \left\|\bm{x}^{(\nu)}(t)-\bbf(\bm{x}(t),\bm{x}^{(1)}(t), \ldots, \bm{x}^{(\nu-1)}(t), W, v)\right\|_2^2dt, \bm{z}\in\bbR^{r_0},
\]

\[
    \check{\bbf}(\bm{z},\check{W}, \check{v}) := \argmin_{\bbf \in \mathcal{F}_2(L, \bm{p}, \tau, F)}\sum_{i=1}^n
    \left\|\bm{x}^{(\nu)}(t_i)-\bbf(\bm{x}(t),\bm{x}^{(1)}(t_i), \ldots, \bm{x}^{(\nu-1)}(t_i), W, v)\right\|_2^2, \bm{z}\in\bbR^{r_0}.
\]
For simplicity, we use $\hat{\bbf}, \tilde{\bbf}, \check{\bbf}$ to represent $\hat{\bbf}(\bm{z},\hat{W}, \hat{v}), \tilde{\bbf}(\bm{z},\tilde{W}, \tilde{v})$ and $\check{\bbf}(\bm{z},\check{W}, \check{v})$.

By the definition of $\hat{\bbf}$ and $\tilde{\bbf}$ we can obtain
\begin{align*} \label {hat tilde inequality}
\int_0^1\|\hat{\bbf}(\widehat{\bm{x}}(t),\widehat{\bm{x}}^{(1)}(t), \ldots, \widehat{\bm{x}}^{(\nu-1)}(t)) - \widehat{\bm{x}}^{(\nu)}(t)\|_2^2dt \leq \int_0^1\|\tilde{\bbf}(\widehat{\bm{x}}(t),\widehat{\bm{x}}^{(1)}(t), \ldots, \widehat{\bm{x}}^{(\nu-1)}(t)) - \widehat{\bm{x}}^{(\nu)}(t)\|_2^2dt\\
\int_0^1\|\tilde{\bbf}(\bm{x}(t),\bm{x}^{(1)}(t), \ldots, \bm{x}^{(\nu-1)}(t)) - \bm{x}^{(\nu)}(t)\|_2^2dt \leq \int_0^1\|\hat{\bbf}(\bm{x}(t),\bm{x}^{(1)}(t), \ldots, \bm{x}^{(\nu-1)}(t)) - \bm{x}^{(\nu)}(t)\|_2^2dt\\
\int_0^1\|\tilde{\bbf}(\bm{x}(t),\bm{x}^{(1)}(t), \ldots, \bm{x}^{(\nu-1)}(t)) - \bm{x}^{(\nu)}(t)\|_2^2dt \leq \int_0^1\|\check{\bbf}(\bm{x}(t),\bm{x}^{(1)}(t), \ldots, \bm{x}^{(\nu-1)}(t)) - \bm{x}^{(\nu)}(t)\|_2^2dt
\end{align*}

Using the above inequalities, we can decompose the error into the following four terms:
\begin{align*}
    & \int_0^1\|\hat{\bbf}(\bm{x}(t), \bm{x}^{(1)}(t), \ldots, \bm{x}^{(\nu-1)}(t))
      - \bbf_{0}(\bm{x}(t), \bm{x}^{(1)}(t), \ldots, \bm{x}^{(\nu-1)})(t)\|_2^2dt\\
    &\lesssim  \int_0^1\|\hat{\bbf}(\bm{x}(t), \bm{x}^{(1)}(t), \ldots, \bm{x}^{(\nu-1)}(t)) - \hat{\bbf}(\widehat{\bm{x}}(t), \widehat{\bm{x}}^{(1)(t)}, \ldots, \widehat{\bm{x}}^{(\nu-1)}(t))\|_2^2dt\\
     &\quad+ \int_0^1\|\widehat{\bm{x}}^{(\nu)}(t) - \bbf_{0}(\bm{x}(t), \bm{x}^{(1)}(t), \ldots, \bm{x}^{(\nu-1)}(t))\|_2^2dt\\
     &\quad + \int_0^1\|\hat{\bbf}(\widehat{\bm{x}}(t), \widehat{\bm{x}}^{(1)}(t), \ldots, \widehat{\bm{x}}^{(\nu-1)}(t))
      - \widehat{\bm{x}}^{(\nu)}(t)\|_2^2dt\\
     &\lesssim  T_1 + T_2 + \int_0^1\|\tilde{\bbf}(\widehat{\bm{x}}(t),\widehat{\bm{x}}^{(1)}(t), \ldots, \widehat{\bm{x}}^{(\nu-1)}(t)) - \widehat{\bm{x}}^{(\nu)}(t)\|_2^2dt\\
     &\lesssim T_1 + T_2\\
     &\quad+  \int_0^1\|\tilde{\bbf}(\bm{x}(t), \bm{x}^{(1)}(t), \ldots, \bm{x}^{(\nu-1)}(t)) - \tilde{\bbf}(\widehat{\bm{x}}(t), \widehat{\bm{x}}^{(1)}(t), \ldots, \widehat{\bm{x}}^{(\nu-1)}(t))\|_2^2dt\\
     &\quad+ \int_0^1\|\widehat{\bm{x}}^{(\nu)}(t) - \bbf_{0}(\bm{x}(t), \bm{x}^{(1)}(t), \ldots, \bm{x}^{(\nu-1)}(t))\|_2^2dt\\
     &\quad + \int_0^1\|\tilde{\bbf}(\bm{x}(t),\bm{x}^{(1)}(t), \ldots, \bm{x}^{(\nu-1)}(t)) - \bm{x}^{(\nu)}(t)\|_2^2dt\\
     &\lesssim T_1 + 2T_2 + T_3\\
     &\quad+ \int_0^1\|\check{\bbf}(\bm{x}(t),\bm{x}^{(1)}(t), \ldots, \bm{x}^{(\nu-1)}(t)) - \bm{x}^{(\nu)}(t)\|_2^2dt\\
     & = T_1 + 2T_2 + T_3 + T_4,
\end{align*}
where
\begin{align*}
T_1 &= \int_0^1\|\hat{\bbf}(\bm{x}(t), \bm{x}^{(1)}(t), \ldots, \bm{x}^{(\nu-1)}(t)) - \hat{\bbf}(\widehat{\bm{x}}(t), \widehat{\bm{x}}^{(1)}(t), \ldots, \widehat{\bm{x}}^{(\nu-1)})(t)\|_2^2dt,\\
T_2 &= \int_0^1\|\widehat{\bm{x}}^{(\nu)}(t) - \bbf_{0}(\bm{x}(t), \bm{x}^{(1)}(t), \ldots, \bm{x}^{(\nu-1)}(t))\|_2^2dt,\\
T_3 &= \int_0^1\|\tilde{\bbf}(\bm{x}(t), \bm{x}^{(1)}(t), \ldots, \bm{x}^{(\nu-1)}(t)) - \tilde{\bbf}(\widehat{\bm{x}}(t), \widehat{\bm{x}}^{(1)}(t), \ldots, \widehat{\bm{x}}^{(\nu-1)}(t))\|_2^2dt,\\
T_4 &= \int_0^1\|\check{\bbf}(\bm{x}(t),\bm{x}^{(1)}(t), \ldots, \bm{x}^{(\nu-1)}(t)) - \bm{x}^{(\nu)}(t)\|_2^2dt.
\end{align*}

The error consists of four parts. $T_1$ and $T_3$ can be viewed as the perturbation error of neural network. $T_2$ is the approximation error from the kernel estimator in stage 1. The last term $T_4$ is the approximation error of neural network. We will analyze these four terms separately.

Suppose $x_j(t)$ is $k$ times differentiable and there is a modified kernel $K_{\nu, q}$. Furthermore, if the sequence $\{s_i\}$ satisfies $\max_i|s_i-s_{i-1}-n^{-1}|=O(n^{-\delta})$ for some $\delta>1$, it follows from Theorem 5 in \cite{Muller1984} that
\begin{equation}\label{T3}
       \int_0^1\|\widehat{\bm{x}}^{(\nu)}(t) - \bbf_{0}(\bm{x}(t), \bm{x}^{(1)}(t), \ldots, \bm{x}^{(\nu-1)}(t))\|_2^2dt = O_P(n^{-2(k-\nu)/(2k+1)}),
\end{equation}
where $\widehat{\bm{x}}^{(\nu)}(t)  = (\widehat{x}_1^{(\nu)}(t),\ldots,\widehat{x}_d^{(\nu)}(t))^{\top}$.

For $T_1$, Let $\bm{x} := (\bm{x}(t), \bm{x}^{(1)}(t), \ldots,\bm{x}^{(\nu-1)}(t))$, $\hat{\bm{x}} := (\widehat{\bm{x}}(t), \widehat{\bm{x}}^{(1)}(t), \ldots, \widehat{\bm{x}}^{(\nu-1)}(t))$, and  $\bm{\delta} := \hat{\bm{x}} - \bm{x} \in\bbR^{r_0}$. As $n \to \infty$, by (\ref{T3}), we know $|\bm{\delta}| \lesssim (\frac{1}{n^{2(k-\nu)/(2k+1)}}, \ldots,\frac{1}{n^{2(k-\nu)/(2k+1)}})$. Meanwhile, define $\hat{\bbf}(\bm{x}, \hat{W}, \hat{v}) = \hat{W}_{L}\sigma_{\hat{\bm{v}}_{L}} \ldots \hat{W}_1 \sigma_{\hat{\bm{v}}_1} \hat{W}_0 \bm{x}$.
Let $\tau_j = \|\hat{W}_j\|_0 + |\hat{\bm{v}}_j|_0$, the number of non-zero entries in the $j$-th layer. Since $\max(\|\hat{W}_j\|_\infty , |\hat{\bm{v}}_j|_\infty) \leq 1$ and the fact $|\sigma_{\hat{\bm{v}}_j}(\bm{x}+\bm{\delta}) - \sigma_{\hat{\bm{v}}_j}(\bm{x})|\leq|\bm{\delta}|$, we can easily get
\begin{align*}
&\|\hat{\bbf}(\bm{x}, \bm{x}^{(1)}, \ldots, \bm{x}^{(\nu-1)}) - \hat{\bbf}(\widehat{\bm{x}}, \widehat{\bm{x}}^{(1)}, \ldots, \widehat{\bm{x}}^{(\nu-1)})\|_\infty^2 \\
=& \|\hat{W}_{L}\sigma_{\hat{\bm{v}}_{L}} \hat{W}_{L-1}\sigma_{\hat{\bm{v}}_{L-1}}\ldots\sigma_{\hat{\bm{v}}_0} \hat{W}_0 \bm{x} - \hat{W}_{L}\sigma_{\hat{\bm{v}}_{L}} \hat{W}_{L-1}\sigma_{\hat{\bm{v}}_{L-1}}\ldots\sigma_{\hat{\bm{v}}_0} \hat{W}_0 (\bm{x} + \bm{\delta})\|_\infty^2\\
\leq& \tau_0\|\hat{W}_{L}\sigma_{\hat{\bm{v}}_{L}}\hat{W}_{L-1}\sigma_{\hat{\bm{v}}_{L-1}}\ldots\sigma_{\hat{\bm{v}}_1} \hat{W}_1 \bm{\delta}\|_\infty^2\\
\leq&\Pi_{j=0}^L \tau_i\|\bm{\delta}\|_\infty^2 \\
\leq& (\frac{\sum_{j=1}^L\tau_j}{L})^L\|\bm{\delta}\|_\infty^2 = \left(\frac{\tau}{L}\right)^L\left\|\bm{\delta}\right\|_\infty^2 = O_p(\left(\frac{\tau}{L}\right)^Ln^{-2(k-\nu)/(2k+1)}),
\end{align*}
which implies $T_1 = O_P(\left(\frac{\tau}{L}\right)^Ln^{-2(k-\nu)/(2k+1)})$.

For $T_2$, by (\ref{T3}), we know $T_2 = O_P(n^{-2(k-\nu)/(2k+1)})$.

For $T_3$, similar to $T_1$, we can show $T_3 = O_P(\left(\frac{\tau}{L}\right)^Ln^{-2(k-\nu)/(2k+1)})$.

For $T_4$, since each true functions $f_{0,j} \in \mathcal{CS}(L_*, \bm{r}, \bm{\tilde{r}}, \bm{\beta},\bm{a}, \bm{b}, \bm{C})$, $j=1, \ldots,d$, let us consider the following estimation problems:
\[
\tilde{f}_j = \argmin_{f \in \mathcal{F}_2(L, \bm{p}, \tau, F)}\int_0^1
    \left\|\bm{x}^{(\nu)}_j(t)-f(\bm{x}(t),\bm{x}^{(1)}(t), \ldots, \bm{x}^{(\nu-1)}(t), W, v)\right\|_2^2dt,
\]
\[
\check{f}_j = \argmin_{f \in \mathcal{F}_2(L, \bm{p}, \tau, F)}\sum_{i=1}^{n}
    \left\|\bm{x}^{(\nu)}_j(t_i)-f(\bm{x}(t_i),\bm{x}^{(1)}(t_i), \ldots, \bm{x}^{(\nu-1)}(t_i), W, v)\right\|_2^2.
\]

By Theorem 2 in \cite{schmidt-hieber}, we get the following inequality
\begin{equation} \label{T4 1}
\int_0^1\|\tilde{f}_j - f_{0,j}\|_2^2dt \leq \int_0^1\|\check{f}_j - f_{0,j}\|_2^2dt  = O\left(\inf_{f \in \mathcal{F}_2(L, \bm{p}, \tau, F)}\|f - f_{0,j}\|_\infty^2\right).
\end{equation}
Now we need to analyze $\inf_{f \in \mathcal{F}_2(L, \bm{p}, \tau, F)}\|f - f_{0,j}\|_\infty^2$. Since $f_{0,j} \in \mathcal{CS}(L_*, \bm{r}, \bm{\tilde{r}}, \bm{\beta},\bm{a}, \bm{b}, \bm{C})$, $j=1, \ldots,d$, by Lemma \ref{lemma 1}, for any $m>0$, there exists a neural network
\[
f_{*,j} \in \mathcal{F}_2(L, (r_0, N,\ldots, N, 1), \tau, \infty),
\]
with $L \asymp m, N \geq 6\eta\max_{i=0,\ldots,L_*}(\beta_i+1)^{\tilde{r}_i} \vee (\tilde{C}_i+1)e^{\tilde{r}_i}, \eta = \max_{i=0,\ldots, L_*}(r_{i+1}(\tilde{r}_i + \ceil{\beta_i})), \tau\lesssim mN,$, such that
\begin{align*}
\| f_{*,j} - f_{0,j}\|_{\infty} &\lesssim \sum_{i=0}^{L_*}(N2^{-m})^{\prod_{l=i+1}^{L_*}\beta_l\wedge1} + (N^{-\frac{\beta_i}{\tilde{r}_i}})^{\prod_{l=i+1}^{L_*}\beta_l\wedge1}\\
&\lesssim \sum_{i=0}^{L_*}(N2^{-m})^{\prod_{l=i+1}^{L_*}\beta_l\wedge1} + N^{-\frac{\beta_i^*}{\tilde{r}_i}}\\
&\lesssim (N2^{-m})^{\prod_{l=1}^{L_*}\beta_l\wedge1} + N^{-\frac{\beta^*}{r^*}},
\end{align*}
where $\beta^*$ and $r^*$ are the intrinsic smoothness and intrinsic dimension defined in Definition 3. This means there exists a sequence of networks $(f_n)_n$ such for all sufficiently large $n$, $\| f_n - f_{0,j}\|_{\infty} \lesssim (N2^{-m})^{\prod_{l=1}^{L_*}\beta_l\wedge1} + N^{-\frac{\beta^*}{r^*}}$ and $f_n \in \mathcal{F}_2(L, \bm{p}, \tau, \infty)$. Next define $\grave{f_j}:=f_n(\|f_{0,j}\|_\infty/\|f_{n}\|_\infty\wedge1) \in \mathcal{F}_2(L, \bm{p}, \tau, F)$ and $\|\grave{f_j} - f_{0,j}\|_{\infty}\lesssim (N2^{-m})^{\prod_{l=1}^{L_*}\beta_l\wedge1} + N^{-\frac{\beta^*}{r^*}}$. Then it follows that $\inf_{f \in \mathcal{F}_2(L, \bm{p}, \tau, F)}\|f - f_{0,j}\|_\infty \lesssim\|\grave{f_j} - f_{0,j}\|_{\infty}\lesssim (N2^{-m})^{\prod_{l=1}^{L_*}\beta_l\wedge1} + N^{-\frac{\beta^*}{r^*}}$.

Since the length, minimum width, active neurons for all $\tilde{f}_j$ has the same order, and $d$ is a fixed constant, we can synchronize the number of hidden layers for all $\tilde{f}_j$ by adding the some additional layers with identity weight matrix. After parallelizing the above $d$ networks $\tilde{f}_j$, we can get the joint neural network $\bbf^* \in \mathcal{F}_2(L, (r_0, N,\ldots, N, d), \tau, F)$ satisfying $F \geq \max(\max_{i=0,\ldots,L^*}C_i, 1)$, with $L \asymp m, N \geq \max_{i=0,\ldots,L_*}(\beta_i+1)^{\tilde{r}_i} \vee (\tilde{C}_i+1)e^{\tilde{r}_i}), \tau\lesssim mN$, such that
\[
\| \bbf^* - \bbf_{0}\|_{\infty} \lesssim (N2^{-m})^{\prod_{l=1}^{L_*}\beta_l\wedge1} + N^{-\frac{\beta^*}{r^*}}.
\]

Combine with (\ref{T4 1}), it holds that
\[
T_4 = O\left((N2^{-m})^{2\prod_{l=1}^{L_*}\beta_l\wedge1} + N^{-\frac{2\beta^*}{r^*}}\right).
\]

Combining the above we get
\[
 \varsigma _{n}= (1+N^L)n^{-2(k-\nu)/(2k+1)} + (N2^{-L})^{2\prod_{l=1}^{L_*}\beta_l\wedge1} + N^{-\frac{2\beta^*}{r^*}}
\]

As a consequence, by the fact that
\begin{align*}
&\int_0^1\|\hat{\bbf}(\widehat{\bm{x}}(t), \widehat{\bm{x}}^{(1)}(t), \ldots, \widehat{\bm{x}}^{(\nu-1)}(t)) - \bbf_{0}(\bm{x}(t), \bm{x}^{(1)}(t), \ldots, \bm{x}^{(\nu-1)}(t))\|_2^2dt \\
& = \int_0^1\|\hat{\bbf}(\widehat{\bm{x}}(t), \widehat{\bm{x}}^{(1)}(t), \ldots, \widehat{\bm{x}}^{(\nu-1)}(t)) - \hat{\bbf}(\bm{x}(t), \bm{x}^{(1)}(t), \ldots, \bm{x}^{(\nu-1)}(t))\\
& \quad + \hat{\bbf}(\bm{x}(t), \bm{x}^{(1)}(t), \ldots, \bm{x}^{(\nu-1)}(t)) - \bbf_{0}(\bm{x}(t), \bm{x}^{(1)}(t), \ldots, \bm{x}^{(\nu-1)}(t))\|_2^2dt\\
& \lesssim T_1 + \int_0^1\|\hat{\bbf}(\bm{x}(t), \bm{x}^{(1)}(t), \ldots, \bm{x}^{(\nu-1)}(t))
      - \bbf_{0}(\bm{x}(t), \bm{x}^{(1)}(t), \ldots, \bm{x}^{(\nu-1)}(t))\|_2^2dt\\
& = O_P(\varsigma_n).
\end{align*}
Combining the above we get the desired result.

\section{Hierarchical Proximal Operator}
\begin{algorithm}
\caption{Hierarchical Proximal Operator}
\label{algo3}
\begin{algorithmic}
\State \textbf{Procedure}: $\textrm{Hier-Prox}(\theta, W_0, \lambda, M)$
\For{ $i \in \{1, \ldots, r_0\}$}
\State Sort the entries of $W_{0,i}$ into $|W_{0,i}^{(1)}|\geq |W_{0,i}^{(2)}|\ldots \geq |W_{0,i}^{(r_1)}|$
\For{ $j \in \{1, \ldots, r_1\}$}
\State Compute $\omega_{j}=\frac{M}{1+jM^2}S_\lambda\left(|\theta_i| + M\sum_{k=1}^j|W_{0,i}^{(k)}|\right)$
\State Find the first $j$ such that $|W_{0,i}^{(j+1)}| \leq \omega_j \leq |W_{0,i}^{(j)}|$
\EndFor
\State $\widetilde{\theta}_i \leftarrow \frac{1}{M}\textrm{sign}(\theta_i)\omega_j$; $\widetilde{W}_{0,i} \leftarrow \textrm{sign}(W_{0,i})\min\{\omega_j, W_{0,i}\}$
\EndFor
\State \textbf{Return}: $(\widetilde{\theta}, \widetilde{W}_0)$
\State \textbf{Note}:
\begin{itemize}
\item $r_1$ is the number of node in the first hidden layer.
\item $S_\lambda(x) = \textrm{sign}(x)\max\{|x|-\lambda, 0\}$.
\item We assume $W_{0,i}^{(0)} = +\infty$ and $W_{0,i}^{(r_0+1)} = 0$.
\end{itemize}
\end{algorithmic}
\end{algorithm}
\end{document}